\pdfoutput=1
\newif\ifarxiv
\arxivtrue

\documentclass[11pt]{article}

\ifarxiv
    \usepackage{EMNLP2023}
\else
    \usepackage[review]{EMNLP2023}
\fi

\usepackage{times}
\usepackage{latexsym}

\usepackage[T1]{fontenc}

\usepackage[utf8]{inputenc}

\usepackage{microtype}

\usepackage{inconsolata}
\usepackage{enumitem}
\usepackage{array}
\usepackage[capitalise]{cleveref}
\usepackage{listings}
\usepackage{graphicx}
\usepackage{longtable}
\usepackage{booktabs}
\definecolor{deepblue}{rgb}{0,0,0.5}
\definecolor{deepred}{rgb}{0.6,0,0}
\definecolor{deepgreen}{rgb}{0,0.5,0}
\definecolor{pink}{RGB}{219, 48, 122}
\definecolor{forestgreen}{RGB}{34,139,34}
\definecolor{goldenrod}{RGB}{218,165,32}
\definecolor{sepia}{RGB}{112,66,20}
\crefname{figure}{Figure}{Figures}
\crefname{table}{Table}{Tables}
\crefname{appendix}{Appendix}{Appendices}
\crefname{section}{Section}{Sections}
\crefname{equation}{Eq.}{Eqs.}

\newcommand\myparagraph[1]{
\vskip 0.05in 
\noindent{\bf {#1}}}
\title{\textsc{chatReport}: Democratizing Sustainability Disclosure Analysis through LLM-based Tools}

\author{Jingwei Ni\textsuperscript{\rm 1,2} 
    Julia Bingler\textsuperscript{\rm 3,4}
    Chiara Colesanti-Senni\textsuperscript{\rm 1}
    Mathias Kraus\textsuperscript{\rm 5} 
    Glen Gostlow\textsuperscript{\rm 1} \\
    {\bf Tobias Schimanski}\textsuperscript{\rm 1} 
    {\bf Dominik Stammbach}\textsuperscript{\rm 2} 
    {\bf Saeid Ashraf Vaghefi}\textsuperscript{\rm 1,6}
    {\bf Qian Wang}\textsuperscript{\rm 1} \\
    {\bf Nicolas Webersinke}\textsuperscript{\rm 5}
    {\bf Tobias Wekhof}\textsuperscript{\rm 1,2} 
    {\bf Tingyu Yu}\textsuperscript{\rm 1}
    {\bf Markus Leippold}\textsuperscript{\rm 1,7} \\
    \textsuperscript{\rm 1}University of Zurich 
    \textsuperscript{\rm 2}ETH Zurich 
    \textsuperscript{\rm 3}University of Oxford 
    \textsuperscript{\rm 4}Council on Economic Policies \\
    \textsuperscript{\rm 5}FAU Erlangen-Nürnberg 
    \textsuperscript{\rm 6}Eawag: Swiss Federal Institute of Aquatic Science \\
    \textsuperscript{\rm 7}Swiss Finance Institute (SFI) \\ 
    \texttt{njingwei@ethz.ch}
    }

\begin{document}
\maketitle

\begin{abstract}
In the face of climate change, are companies really taking substantial steps toward more sustainable operations? A comprehensive answer lies in the dense, information-rich landscape of corporate sustainability reports.
However, the sheer volume and complexity of these reports make human analysis very costly.
Therefore, only a few entities worldwide have the resources to analyze these reports at scale, which leads to a lack of transparency in sustainability reporting.
Empowering stakeholders with LLM-based automatic analysis tools can be a promising way to democratize sustainability report analysis.
However, developing such tools is challenging due to (1) the hallucination of LLMs and (2) the inefficiency of bringing domain experts into the AI development loop. 
In this paper, we introduce \textsc{chatReport}, a novel LLM-based system to automate the analysis of corporate sustainability reports, addressing existing challenges by (1) making the answers traceable to reduce the harm of hallucination and (2) actively involving domain experts in the development loop. 
We make our methodology, annotated datasets, and generated analyses of 1015 reports publicly available.\footnote{Web app: https://reports.chatclimate.ai/ Demo video: https://www.youtube.com/watch?v=Q5AzaKzPE4M\&t=15s}\footnote{https://github.com/EdisonNi-hku/chatreport}
\end{abstract}

\section{Introduction}
\begin{figure*}[ht!]
	\centering
	\includegraphics[width=1\textwidth]{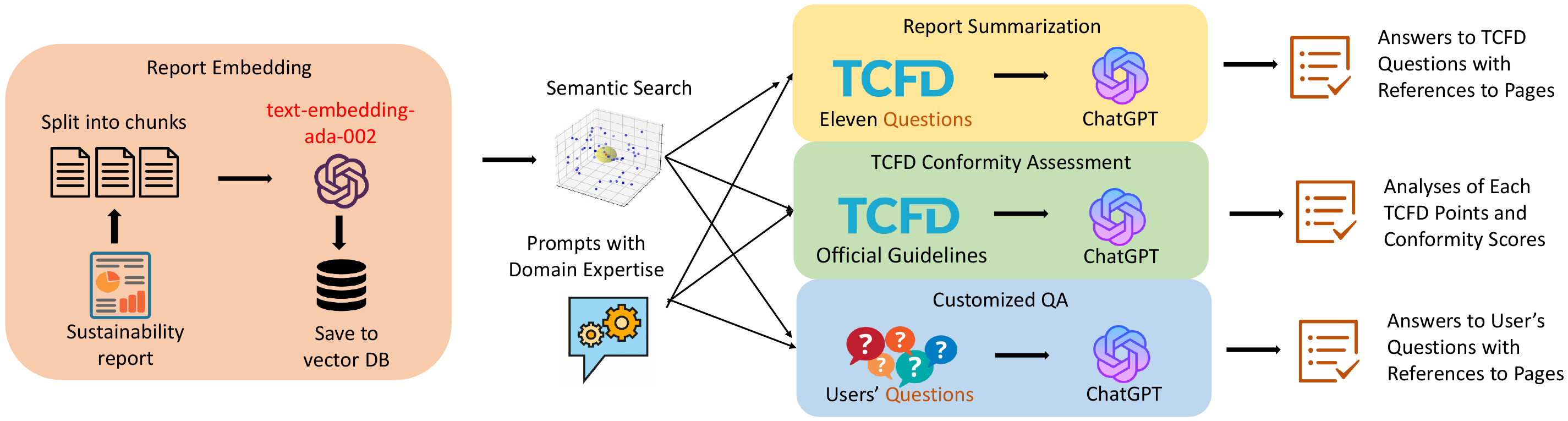}
	\caption{\textsc{chatReport} Pipeline 
 }
	\label{fig1:chatreport}
\end{figure*}
As climate change becomes an increasingly urgent issue, sustainability is becoming a key global concern, necessitating transparent public oversight of corporate sustainability practices. However, the substantial length of sustainability reports (often more than 70 pages) makes it challenging for the majority of stakeholders (including investors, policymakers, and the general public) to digest and analyze them. At the same time, relying on third-party rating agencies is not always a solution, as their services can be expensive, lack transparency, and vary due to differing criteria for evaluating sustainability performance \citep{berg2022aggregate}.

In light of these challenges, automated and transparent approaches are essential to improving accessibility, efficiency, and accuracy when analyzing corporate sustainability reports. 

Large Language Models (LLMs) \citep[][inter alia]{brown2020language,ouyang2022training,touvron2023llama,gpt4techreport,touvron2023llama2} have revolutionized Natural Language Processing (NLP), enabling advancements in automated reasoning, understanding, and generation of text. Such advances can assist in conducting comprehensive analyses of corporate sustainability reports automatically. However, to develop such an LLM-based system, there are two major challenges: LLMs (1) may hallucinate in their outputs \citep{Ji_2023}, and 
(2) have no expertise in sustainability report analysis. Furthermore, there exists no framework which would actively involve domain experts in the prompt development loop, injecting domain expertise into the prompts.

In this paper, we propose \textsc{chatReport}, a system that automatically analyzes sustainability reports based on the TCFD\footnote{We choose TCFD instead of other disclosure guidelines because it is widely adopted and investor-friendly. \cref{appendix:tcfd} covers an introduction for TCFD.} (Task Force on Climate-related Financial Disclosures) recommendations. It computes the reports' conformity score to TCFD guidelines, proposing the first automatic metric for disclosure quality benchmarking. \textsc{chatReport} also supports customized analysis with user question answering. To reduce hallucination, we ground the analytical prompts with retrieved information from the target report, and further make the answers traceable to help users identify hallucinations. To actively bring domain experts into the development loop, we design an automatic prompt engineering algorithm that transfers experts' feedback on specific outputs to general analysis guidelines, which can be injected into our prompt template for future analysis.

Furthermore, we conduct a rigorous human evaluation to analyze the system's 
hallucination rate quantitatively. We find that the system achieves an admirable hallucination-free rate. For those hallucinated cases, it is easy for users to identify them because the system always (1) refers to relevant sources and pages; and (2) answers questions in an extractive manner, making it convenient to identify evidence sentences by keyword search. Moreover, we achieve a moderate inter-annotator agreement on annotating hallucination (Cohen's Kappa of 0.54), further illustrating that the discrepancies between answers and references are easy to identify. Our human evaluation results in an annotated dataset of LLM outputs with attributions, which may contribute to other domains (e.g., LLM attribution verification \citep{yue2023automatic} to check the supportiveness of cited sources for the answer.).
Our contributions include:
\begin{enumerate}[itemsep=0pt,topsep=1pt]
    \item We introduce \textsc{chatReport}, a novel system that automatically analyzes sustainability reporting along different dimensions.
    \item We develop an efficient framework to actively involve domain experts in AI tool development, which may potentially benefit all interdisciplinary research.
    \item We conduct a human evaluation on \textsc{chatReport}'s hallucination and attribution. The resulting dataset contributes to automatic attribution verification.
\end{enumerate}

\section{Related Work and Background}

\myparagraph{NLP for Climate Change}
NLP technologies have been employed in various areas, including meta-analyses in climate science \cite{callaghan2021machine}, or for financial climate disclosure analyses \cite{bingler2022cheap,luccioni2020analyzing}, detecting stance in media about global warming \cite{luo-etal-2020-detecting}, detecting environmental claims \cite{stammbach2022dataset}, and climate claims fact-checking \cite{diggelmann2020climate,webersinke2022climatebert}. More recently, \citet{vaghefi2023chatclimate} introduced \textsc{chatClimate}, a chatbot based on the latest IPCC Assessment Report. By leveraging NLP, researchers aim to extract valuable insights from textual data related to climate change to advance research, decision-making, and public engagement.

\myparagraph{Large Language Models} 
LLMs have emerged as the de-facto standard in recent years \citep{brown2020language,ouyang2022training,chowdhery2022palm,touvron2023llama,anil2023palm,gpt4techreport,touvron2023llama2}. Instruction fine-tuned models, such as ChatGPT \cite{instructgpt} and GPT-4 \cite{gpt4techreport}, have showcased their potential on comprehensive prompt-based AI applications \citep{shen2023hugginggpt,schick2023toolformer}. Some strong LLMs can even be a cheap and reliable proxy for human preference, evaluating the quality of generated texts \citep{chiang2023vicuna, kocmiLargeLanguageModels2023, zhengJudgingLLMasajudgeMTBench2023}.

However, hallucination still remains a major limitation of the SOTA LLMs \cite{Ji_2023}. Related work has proposed initial efforts to (1) better align LLMs \citep{zhou2023lima}; and (2) fight false attribution from LLM-based search engine \citep{liu2023evaluating,yue2023automatic} and LLM-generated misinformation \citep{peng2023check,liSelfCheckerPlugandPlayModules2023}. These efforts suggest potential ways to mitigate LLM hallucinations, but still left it as an open research question.

\myparagraph{Utilizing Experts' Feedback} 
Involving a human in the loop has a long history in machine learning and NLP. However, previous work mainly focuses on active learning \citep{Raghavan2006ActiveLW,wu2021active} and using human feedback to improve specific outputs \citep{elgohary-etal-2020-speak,tandon2021interscript}. In this work, we propose a novel prompting-based approach to automatically improve general prompts using experts' feedback on specific outputs, which actively brings human experts into the prompt engineering loop.

\section{\textsc{chatReport}}

\subsection{Pipeline}

The pipeline of \textsc{chatReport} is illustrated in \cref{fig1:chatreport}. Given a sustainability report, \textsc{chatReport} analyzes it with the following four modules.

\myparagraph{Report Embedding (RE)}
To address the limited context window, the RE module first splits the report into text chunks, which are then transformed into a vector space representation for future reference and semantic searching. We have domain experts transfer TCFD recommendations to queries for retrieval (details in \cref{appendix:questions}).

\myparagraph{Report Summarization (RS)}
To assist in efficiently reading the report, the RS module summarizes it based on TCFD's eleven recommended aspects that companies are asked to describe. Given each TCFD recommendation, the RS module first retrieves the relevant part from the report using our carefully designed query. Then it prompts the LLM to summarize the report's disclosure on that TCFD recommendation, with the retrieved part (from the RE module) and the company's basic information as context. Prompt templates for this module can be found in \cref{appendix:qa_or_summary}.

\myparagraph{TCFD Conformity Assessment (TCA)}
In addition to the recommendations, TCFD also provides detailed disclosure guidelines for each recommendation, which specify the type and granularity of information that companies need to disclose in their report. To evaluate the reports' conformity to TCFD guidelines, we design the TCA module to analyze to which extent the report follows TCFD guidelines: for each TCFD recommendation, the TCA module takes in relative contexts from the RE module. It then evaluates it against the respective TCFD guidelines, generating an analysis paragraph and a TCFD conformity score from 0 to 100. The prompt template for this model can be found in \cref{appendix:conformity_prompt}.

By explicitly defining the scoring criteria and providing clear instructions, we aim to minimize potential biases and enhance the reliability of the evaluation process. However, it is essential to acknowledge that the LLM-generated scores might be far from perfect \citep{zhengJudgingLLMasajudgeMTBench2023}.  
We believe that the scoring strategy implemented in our study represents a valid and valuable first step toward leveraging AI-based and automated methods for rating sustainability reports. We encourage future research and collaborative efforts to refine and improve this scoring strategy, considering alternative perspectives by including additional data sources and engaging a broader range of stakeholders.\footnote{We recall that our TCFD conformity score is not a rating or assessment of actual actions or commitments made by companies to address climate change. Instead, it measures the extent to which companies disclose relevant climate-related information in their financial reports.}

\myparagraph{Customized Question Answering (CQA)}
Beyond the analytical structure provided by our framework, we enable users to conduct a personalized analysis by posing customized questions. Our prompt template takes in the user's question and the retrieved relevant contexts which are queried by the question itself (using the RE module). Then, the CQA module makes an LLM call to answer the question. The CQA module's prompt template is almost the same as the RS module's question-answering prompt template, but with slightly different responding guidelines to deal with the noisier scenario where the questions are customized by the users (see details in \cref{appendix:guidelines}).

\subsection{Implementation Details}
We use ChatGPT as the base LLM to conduct experiments and analysis in this paper. 
We use LangChain\footnote{https://python.langchain.com/} to manage OpenAI API calls and vector-database retriever. We use OpenAI's text-embedding-ada-002 for text chunk embedding. Empirically, we find that splitting reports into chunks of $500$ characters (with an overlap of $20$ characters between chunks) results in the best retrieval performance. We usually retrieve the top $20$ related chunks from the RE module. If the prompt becomes too long (e.g., more than 4000 tokens) after inserting the retrieved chunks, we gradually remove the least relevant chunks until the prompt is suitable for the context window. We set the temperature to 0 for all LLM calls and reuse a static vector database for each report.

\subsection{Answer Traceability}
To reduce hallucinations and improve interpretability, we attach source numbers to retrieved chunks and prompt the LLM to provide its attribution (i.e., the chunks it refers to when summarizing information about TCFD recommendations and answering users' questions). With the references attached, human experts can efficiently check whether the model produces misinformation. In \cref{sec:traceability}, we quantitatively analyze the system's answer traceability on a sampled set of outputs.

\subsection{Expert-Involved Prompt Development}
Prompt development is the critical part of \textsc{chatReport} to make sure the outputs (1) contain granular details that stakeholders care about; (2) are formulated in an honest and traceable way; and (3) demonstrate awareness of potential cheap talk and greenwashing. To accomplish this, it is crucial for our domain experts to actively participate in prompt development. We first write several prompt templates, choosing one of them based on domain experts' feedback on outputs. Then we empower domain experts with an LLM-based automatic prompt engineering tool, enabling them to fine-tune the prompts' specifics autonomously, without the help of human prompt engineers. Details are described below:

\myparagraph{Prompt Template Selection: Question Answering or Summarization?} 
There are multiple ways to prompt an LLM to summarize a report's disclosure regarding TCFD recommendations. One is to directly copy the original TCFD recommendation and prompt the LLM to summarize it (e.g., In governance, the company must describe the board's oversight of climate-related risks and opportunities). Another is to first transfer the recommendation to a question about the report (see \ref{appendix:questions} for question-transformation details), then prompt the LLM to answer the question (e.g., How does the company's board oversee climate-related risks and opportunities?). Our prompts for both scenarios are disclosed in Appendix \ref{appendix:qa_or_summary}. We evaluate the prompt templates with experts involved, where the expert's feedback shows that question answering outperforms disclosure summarization (one example is shown in \cref{tab:prompt_dev}).

\begin{figure}[t!]
	\centering
	\includegraphics[width=1\linewidth]{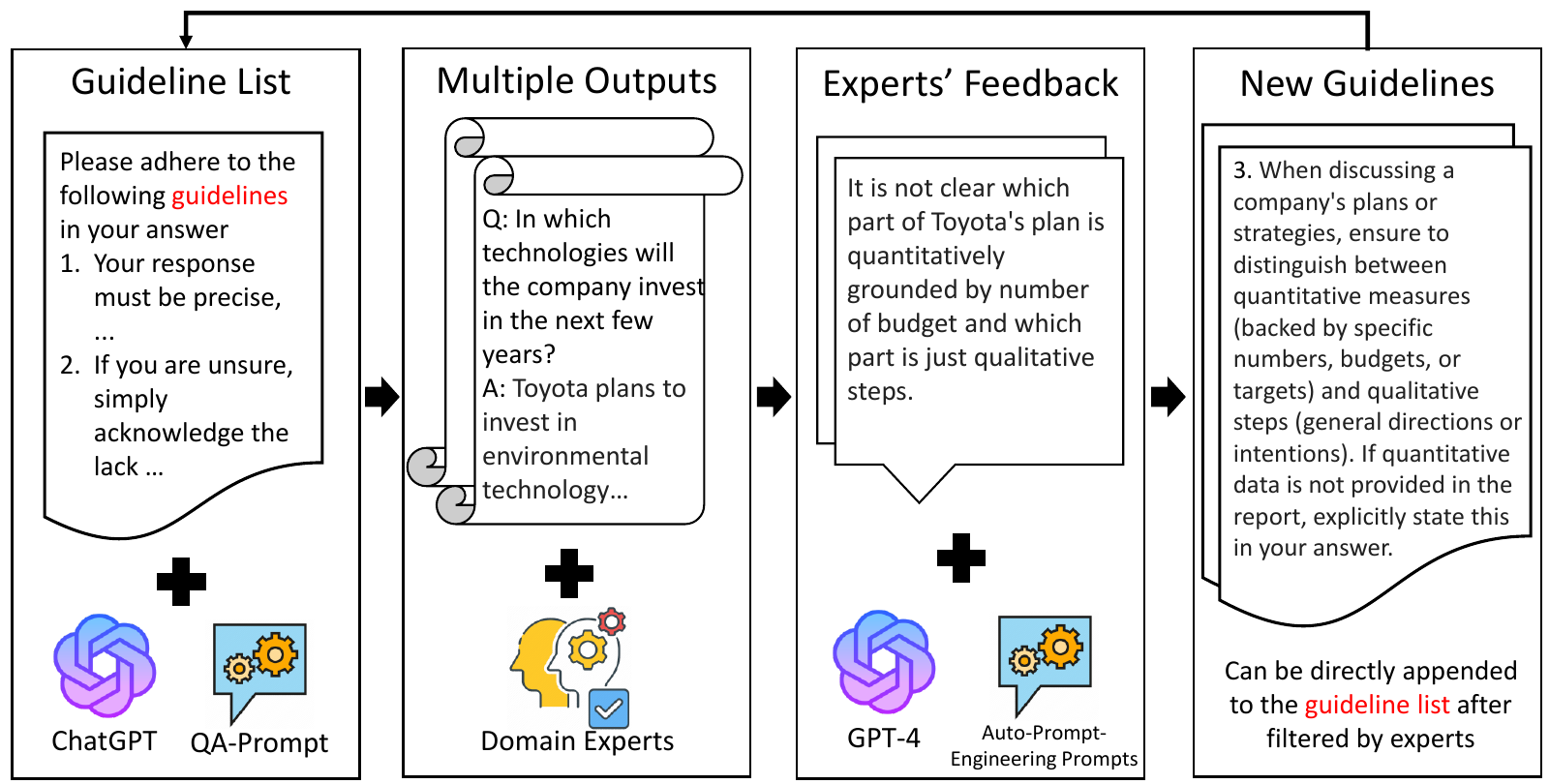}
	\caption{Automatic prompt engineering pipeline. 
 }
	\label{fig:auto_prompt}
\vspace{-1.5em}
\end{figure}

\myparagraph{Automatic Prompt Engineering} Without granular adjustment on prompts, ChatGPT's analysis of a sustainability report differs a lot from a human expert's. For example, ChatGPT tends to flatter the user (due to its instruction-following nature), answering with optimism prior and becoming less critical of the possible cheap talk and greenwashing in the report. ChatGPT also tends to be wordy, including irrelevant or even hallucinated information in its response. Moreover, analysts usually expect critical information from a good summarization corresponding to TCFD recommendations, for example, the quantifiability and verifiability of the disclosure. However, ChatGPT fails to include such information because it is not explicitly stated in the TCFD recommendations. Analysts may even expect specific implicit information for each recommendation. To better incorporate these comprehensive, specific, and granular requirements in prompting, we design an automatic prompt engineering tool so that the domain experts can efficiently transfer their feedback on specific outputs to general analysis guidelines which can be used to improve the prompts.

The workflow of automatic prompt engineering is illustrated in \cref{fig:auto_prompt}. The domain experts first suggest improvements for specific answers. Then we prompt ChatGPT to transform the feedback into guidelines that can be used to guide future TCFD question answering. In our prompt template, there is a list of guidelines that the LLM needs to adhere to in its answer. The generated guidelines are then appended to this list to improve the prompts. We started with a guideline list containing general guidelines for honest question answering:
\begin{lstlisting}[frame=single, basicstyle=\ttfamily\scriptsize, xleftmargin=0pt, breaklines, numbers=none]
Please adhere to the following guidelines in your answer
1. Your response must be precise, thorough, and grounded on specific extracts from the report to verify its authenticity.
2. If you are unsure, simply acknowledge the lack of knowledge, rather than fabricating an answer.
\end{lstlisting}

Then we develop new guidelines based on answers generated with this guideline list. Finally, we select five general guidelines for all question answering and one specific guideline for each TCFD recommendation. The general and specific guidelines can be found in \cref{appendix:guidelines}. Prompts of automatic prompt engineering can be found in \cref{appendix:auto_prompt}.

\subsection{Feedback Collection}
We regard \textsc{chatReport} as an ongoing learning system instead of a static analysis tool. Besides our domain experts, we also want to engage our users in the development and learning loop. We will collect users' feedback on TCFD disclosure summarization and TCFD conformity analysis. Such feedback can either be used for both prompt improvements using our automatic prompt engineering method or be saved for memory and reflection for future refinements \citep{tandonLearningRepairRepairing2022}.

\section{Usages}
We collected 9781 sustainability reports spanning 2010 to 2022 (fiscal years). Most of the reports are companies that are traded on the NASDAQ and NYSE. We find that the number of pages in corporate sustainability reports has slightly increased over recent years: in the fiscal year 2017, the mean length of the report has been at 59 pages. In 2021, this number increased to 70 pages, illustrating the increasing effort required by analyzing the reports manually.

\myparagraph{TCFD Conformity Analysis}
\begin{figure}[t!]
	\centering
	\includegraphics[width=1\linewidth]{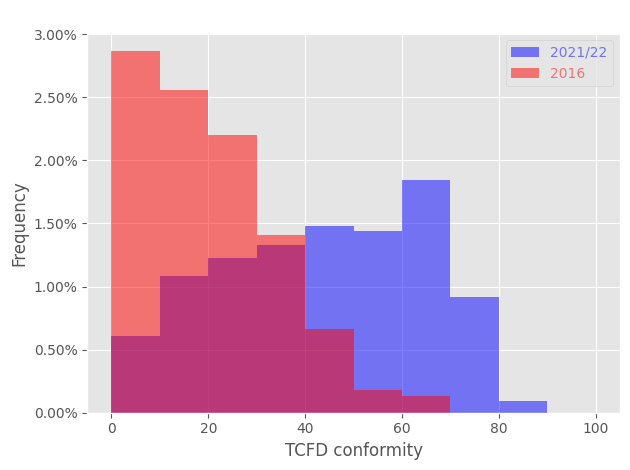}
	\caption{Density plot of the TCFD conformity for an arbitrary sample of corporate sustainability reports of companies listed on NYSE, sampled from 2016 and 2021/22.}
	\label{fig:Conformity}
\end{figure}
Using the RS and TCA modules, we summarize TCFD disclosures and compute TCFD conformity scores for 1015 sustainability reports. Among these reports, 777 are from 2021 and 2022, while 227 are from 2016. \cref{fig:Conformity} illustrates the distribution of these scores for the two sample sets. Our findings indicate a significant impact of the TCFD recommendations on the average TCFD conformity, suggesting that companies embrace these guidelines. However, it is essential to note that TCFD conformity does not necessarily reflect the genuine commitment of companies toward their climate mitigation goals. It is necessary to consider the possibility of ``cheap talk," where firms may make superficial claims without substantial actions to address climate-related issues \cite{bingler2022initiatives}. In \cref{appendix:conformity_examples}, we showcase TCFD conformity analyses on sustainability reports of JP Morgan Chase, Shell and UBS in detail to illustrate the analytic usage of \textsc{chatReport}.

\myparagraph{Customized Analysis}
The CQA module allows users to customize their analysis through question-answering. \cref{appendix:customized_qa} provides some illustrative examples of valuable analytic questions. Posing these questions allows us to gain valuable insights from the sustainability reports beyond the TCFD requirements summarized by the RS modules.

\section{Hallucination Analysis}

\begin{table}[t]
\small
\centering
\begin{tabular}{lccc}
\hline
  \textbf{Backbone}         & \textbf{Content}   & \textbf{Source}   & $\mathbf{R1/R2/RL}$  
  \\ \hline
ChatGPT  & $\mathbf{83.63}$          & $\mathbf{75.00}$  &  $69.89/35.12/51.48$ 
\\ 
GPT-4  & $69.09$          & $72.37$  &  $\mathbf{85.20/50.31/61.50}$ 
\\ \hline
\end{tabular}
\caption{The \textbf{Content} column shows the hallucination-free rate on the content dimension. The \textbf{Source} column shows the hallucination-free rate on the source dimension when the answer is not hallucinated in content. \textbf{R1/R2/RL} shows the ROUGE-X precision scores using the retrieved report content as references. ChatGPT results are obtained on June 28th, 2023. GPT-4 results are obtained on July 6th, 2023.
}\label{tab:hallucination}

\end{table}
We conduct a human evaluation to assess the frequency and degree of hallucinations in \textsc{chatReport}'s output when answering questions.\footnote{It is important to note that we analyze the answers' honesty instead of quality.} Hallucination is evaluated along two dimensions: (1) \textbf{Content}: An answer is not hallucinated if all its covered information is supported by the report. All answers that are not fully supported (e.g., extrapolation or partial support) are considered hallucinated on the content dimension. (2) \textbf{Source}: An answer is not hallucinated on the source dimension only when the model honestly reports its references and the content is not hallucinated; otherwise, the answer is hallucinated on the source dimension (we use binary annotation inspired by \citep{krishna-etal-2023-longeval}).

We randomly sampled 10 sustainability reports (110 TCFD question-answering pairs in total) for human evaluation (sampling details in \cref{appendix:sample_reports}). We have two different annotators to annotate each answer. If there is a disagreement on labeling, we assign a third annotator to make the decision. We conduct human evaluations on both ChatGPT and GPT-4 as the backbone LLM. 

We surprisingly find that despite our strict annotation standard, \textsc{chatReport} reaches a satisfactory hallucination-free rate. With ChatGPT, it honestly conveys information from the report \textbf{83.63\%} of the time, considering the \textbf{51.5\%} average hallucination-free rate of existing generative search engines reported by \citet{liu2023evaluating}.\footnote{We quote \citeposs{liu2023evaluating} result for reference. These numbers are not fully comparable because of the differences in task and data. See more in \cref{appendix:comparison}.} Further findings and discussion are presented in the following subsections. 

\subsection{Answer Traceability} \label{sec:traceability}
We find that \textsc{chatReport} follows our instructions well by answering questions through copying or close paraphrasing. \cref{tab:hallucination} shows that the answers achieve a high ROUGE precision score against the report content no matter with which backbone LLM, illustrating that the answers tend to adhere to the reports' original utterances. This makes the outputs easy to trace using a simple keyword search. If a piece of information is not entailed by its evidence sentence, we mark it as hallucinated. An example can be found in \cref{appendix:human_eval_example}.

\subsection{How Does \textsc{chatReport} Hallucinate?} 
Most of the hallucinations on the content dimension lie in extrapolating reference chunks. Here is an example where the answer falsely concatenates two separate chunks:
\begin{lstlisting}[frame=single, basicstyle=\ttfamily\scriptsize, xleftmargin=0pt, breaklines, numbers=none, escapeinside={(*@}{@*)}]
(*@\textbf{Retrieved chunks in a prompt:}@*)
Content: ... Assurant may incur additional costs associated with (*@\underline{tracking}@*)
Source: 174

Content: (*@\underline{climate hazards}@*). Own Operations: In addition to those noted in ...
Source: 186

... (more chunks and their source numbers)

(*@\textbf{LLM Answer:}@*) 
... Risks include additional costs associated with (*@\underline{tracking climate hazards}@*), declining property values due to sea-level rise...
\end{lstlisting}
Although we explicitly told the model that the retrieved chunks might contain incomplete sentences at both ends and the chunks are delineated by source numbers and new lines, the LLM occasionally falsely concatenates two chunks or makes erroneous extrapolations based on incomplete sentences. We leave the mitigation and automatic detection of such hallucinations to later versions of \textsc{chatReport} and future work.

\subsection{Which Backbone LLM is More Suitable?}
We surprisingly find that ChatGPT outperforms GPT-4 by a large margin in answer honesty. This is because GPT-4 tends to summarize information at a higher level and make unnecessary inferences when answering the questions, which leads to more hallucination in detail. Since we label an answer hallucinated even if there is only a minor error, many GPT-4 answers are labeled as hallucinated. A comparison between GPT-4 and ChatGPT answers is showcased in \cref{appendix:answer_comparison}.

It is also harder to identify GPT-4's hallucination than ChatGPT. When annotating hallucinations, we reach a Cohen's Kappa of $\mathbf{0.54}$ for ChatGPT but only $\mathbf{0.21}$ for GPT-4. Sometimes, the highly abstractive and paraphrased nature of GPT-4 outputs makes it hard even for our expert annotators to identify hallucinations (though GPT-4 uses more utterances from the original reports as illustrated by the ROUGE precision scores). Therefore, we use ChatGPT as the backbone LLM for \textsc{chatReport}

\subsection{The Annotated Dataset}
Our human evaluation for hallucination results in an expert-annotated dataset with labels of whether the answer is fully supported by the references. Human evaluation of how attributable an answer is to its reference is expensive and time-consuming, future work may study how to automize this evaluation process \citep{yue2023automatic} and benchmark the algorithm using our dataset.

\section{Conclusion and Future Work}
We propose \textsc{chatReport} for automatic sustainability report analysis and demonstrating its potential applications and implications. Our prompt development loop and annotated datasets about hallucination could positively transfer to other NLP and interdisciplinary research. \textsc{chatReport} is an open-sourced ongoing project. Our future work will focus on (1) enhancing the retrieval module to provide more accurate contexts for generation, (2) developing automatic attribution-checking tools to fight hallucination, and (3) migrating from OpenAI models to local LLM for more controllable output.

\section*{Ethical Considerations}
\myparagraph{Generate False Information:} Model hallucination is still a significant unresolved problem in NLP. \textsc{chatReport} also generates hallucinations and requires some manual efforts to trace the answer. Moreover, due to the imperfect retrieval module, \textsc{chatReport} may ignore some relevant information. To avoid causing misinformation and disinformation, on one hand, we disclaim on our website that \textsc{chatReport}'s outputs can only be used as references, and cannot be cited as evidence or factual claims. On the other hand, we are experimenting with different approaches to make the outputs more accurate and will release better versions in the future.
\myparagraph{Bias towards Firm Perspective:} A limitation of our approach is the inherent bias towards the firm's perspective in the extracted information from corporate sustainability reports. As \textsc{chatReport} relies solely on the provided information as reported by the firm, it may struggle to provide unbiased and critical responses to certain questions. To mitigate this limitation, we will explore methods to incorporate external perspectives and independent sources of information in future work. This can be achieved by integrating data from third-party assessments, public opinion surveys, or expert evaluations. By incorporating a broader range of perspectives and data inputs, the analysis can provide a more comprehensive and balanced understanding of corporate sustainability performance.
\myparagraph{Changing Behavior of OpenAI Models:} OpenAI continues to update their model, leading to a changing performance of \textsc{chatReport}'s backbone model \citep{chen2023chatgpts}. This may lead to less or more hallucination rates than we reported. In future work, we will substitute the OpenAI closed-source models with our own LLM checkpoints, making the system more controllable and reproducible. 
\myparagraph{Human annotation:} All human annotators are co-authors of this paper, including climate and NLP researchers who have full knowledge about the context and utility of the collected data. We adhered strictly to ethical guidelines, respecting the dignity, rights, safety, and well-being of all participants. There are no data privacy issues or bias against certain demographics with regard to the annotated data.
\myparagraph{License of the Tool:} We use Apache License 2.0 to enable all stakeholders to use and adapt the Tool.

\section*{Broader Implications}

\myparagraph{Supporting Stakeholder Decision-making:} 
Stakeholders, including investors, customers, employees, and regulatory bodies, heavily rely on corporate sustainability reports to make informed decisions. The automated analysis provided by the framework empowers stakeholders with valuable insights into a company's sustainability performance. Investors can use the extracted indicators to assess the environmental, social, and governance (ESG) risks and opportunities 
Customers can make more sustainable choices by considering a company's sustainability practices. Employees can evaluate a company's commitment to social and environmental responsibility. Regulators can use the analysis results to monitor compliance with sustainability regulations.

\myparagraph{LLMs Disruptive Potential for the Rating Industry: }  
Empowering all stakeholders with an automated analysis framework could significantly diminish the need to rely on rating agencies for sustainability report assessments. 
This shift in power from rating agencies to the general public and investors can potentially disrupt rating agencies' business models and challenge their long-standing dominance in sustainability reporting analysis: 
Rating agencies might start to focus on critical assessments of the information disclosed by companies, and provide external analyses of the strategies, rather than summarizing the firm's information. 

\section*{Acknowledgements} This paper has received funding from the Swiss
National Science Foundation (SNSF) under the
project `How sustainable is sustainable finance? Impact evaluation and automated greenwashing detection' (Grant Agreement No. 100018\_207800).

\bibliography{anthology,custom}
\bibliographystyle{acl_natbib}

\appendix

\section{Question Answering vs Summarization Prompts} \label{appendix:qa_or_summary}
Our two forms of prompt templates for the report summarization module are presented in this section. The content embraced by "\{\}" are generic variables to be replaced by corresponding contents. For \{basic\_info\}. we provide the company's name, location, and sector, which are also retrieved by LLMs. For \{guidelines\}, we use the guideline list developed by our experts (details in Appendix \ref{appendix:guidelines}). For \{retrieved\_chunks\_with\_source\}, we append each retrieved chunk with its chunk and page IDs for reference. For \{A\_TCFD\_recommendation\}, we use the original TCFD recommendations. For \{question\_regarding\_a\_TCFD\_recommendation\}, our experts rewrite each TCFD recommendation into a question form (details in Appendix \ref{appendix:questions}).

\subsection{Prompt for Question Answering}
\begin{lstlisting}[frame=single, basicstyle=\ttfamily\scriptsize, xleftmargin=0pt, numbers=none]
As a senior equity analyst with expertise in climate science evaluating a company's sustainability report, you are presented with the following background information:

{basic_info}

With the above information and the following extracted components (which may have incomplete sentences at the beginnings and the ends) of the sustainability report at hand, please respond to the posed question, ensuring to reference the relevant parts ("SOURCES").
Format your answer in JSON format with the two keys: ANSWER (this should contain your answer string without sources), and SOURCES (this should be a list of the source numbers that were referenced in your answer).

QUESTION: {question_regarding_a_TCFD_recommendation}
=========
{retrieved_chunks_with_source}
=========

Please adhere to the following guidelines in your answer:
{guidelines}

Your FINAL_ANSWER in JSON (ensure there's no format error):
\end{lstlisting}

\subsection{Prompt for Summarization}
\begin{lstlisting}[frame=single, basicstyle=\ttfamily\scriptsize, xleftmargin=0pt, numbers=none]
Your task is to analyze and summarize any disclosures related to the following <CRITICAL_ELEMENT> in a company's sustainability report:

<CRITICAL_ELEMENT>: {A_TCFD_recommendation}

Provided below is some basic information about the company under evaluation:

{basic_info}

In addition to the above, the following extracted sections (which may have incomplete sentences at the beginnings and the ends) of the sustainability report have been made available to you for review:

{retrieved_chunks_with_source}

Your task is to summarize the company's disclosure of the aforementioned <CRITICAL_ELEMENT>, based on the information presented in these extracts. Please adhere to the following guidelines in your summary:
{guidelines}

Your summarization should be formatted in JSON with two keys:
1. SUMMARY: This should contain your summary without source references.
2. SOURCES: This should be a list of the source numbers that were referenced in your summary.

Your FINAL_ANSWER in JSON (ensure there's no format error):
\end{lstlisting}

\begin{table*}[t]
\small
\centering
\resizebox{\textwidth}{!}{
\begin{tabular}{>{\centering\arraybackslash}m{0.1\textwidth}m{0.7\textwidth}m{0.2\textwidth}}
\hline
Setting & Example Outputs & Expert's Comment \\ \hline

Question Answering or Summarization? & \textit{\textbf{Summarization:}} Toyota's Board of Directors oversees climate-related risks and opportunities by \textcolor{forestgreen}{monitoring progress towards targets, assessing risks multiple times a year, and formulating measures to solve climate-related issues.} \textcolor{goldenrod}{The committees} are responsible for assessing and managing climate-related risks and opportunities. \textcolor{blue}{Reporting to the Board of Directors occurs every six months or when an important event arises.}

\textit{\textbf{Question Answering:}} The Board of Directors oversees climate-related risks and opportunities by \textcolor{forestgreen}{deliberating and supervising strategies, major action plans, and business plans, monitoring progress toward qualitative and quantitative targets addressing climate issues, and considering the financial impact of risks and opportunities.} \textcolor{goldenrod}{The Environmental Product Design Assessment Committee and Production Environment Committee are the major bodies} in assessing and managing climate-related risks and opportunities, \textcolor{blue}{reporting to the Board of Directors every six months or when an important event arises.}& Both outputs include information on \textcolor{forestgreen}{the board's oversights on risks and opportunities}, \textcolor{goldenrod}{the responsible committees}, and \textcolor{blue}{the reporting mechanism}. However, question-answering provides more detailed and granular information than summarization. \\ \hline

Answer Length & \textit{\textbf{No length limitation:}} \textcolor{forestgreen}{Toyota uses multiple metrics} to manage climate-related risks and opportunities, including CO2 emissions and other elements related to climate change. The metrics are \textcolor{goldenrod}{integrated into the Toyota Environmental Challenge 2050, 2030 Milestone, and Seventh Toyota Environmental Action Plan}. Toyota \textcolor{goldenrod}{aims to achieve carbon neutrality by 2050} with three zero challenges. \textcolor{magenta}{The company promotes electrification and renewable energy in production processes and takes adaptation measures for natural disasters. The risks relating to climate change are identified and assessed based on the TGRS, and important risks and opportunities are reported to the Board of Directors}. However, the report \textcolor{blue}{lacks specific information on the targets and the company's performance against them}.

\textit{\textbf{50 words:}} \textcolor{forestgreen}{Toyota uses multiple metrics} to manage climate-related risks and opportunities, including CO2 emissions, energy, water, resource recycling, and biodiversity. The metrics are \textcolor{goldenrod}{integrated into the Toyota Environmental Challenge 2050, 2030 Milestone, and Seventh Toyota Environmental Action Plan}. Toyota \textcolor{goldenrod}{aims to achieve carbon neutrality by 2050} with three zero challenges. The report does \textcolor{blue}{not provide sufficient information to evaluate the organization's performance against these targets}. & Both answers cover the key points about Toyota’s \textcolor{forestgreen}{metrics}, \textcolor{goldenrod}{targets}, and \textcolor{blue}{lacked information}. Longer responses tend to \textcolor{magenta}{over-include details that are not directly asked for and are not necessarily more informative}, which may exaggerate the richness of disclosed information. \\ \hline

With or without guidelines on critical thinking & \textit{\textbf{Without:}} Toyota has identified risks and opportunities associated with climate change that may impact its businesses, strategy, and financial planning. The risks are classified into \textcolor{forestgreen}{physical} and \textcolor{goldenrod}{transition risks}, including \textcolor{goldenrod}{tightening regulations for fuel efficiency and zero-emission vehicles, expansion of carbon pricing}, and \textcolor{forestgreen}{increase in frequency and severity of natural disasters}. Toyota aims to \textcolor{blue}{achieve carbon neutrality by 2050} and has set multiple metrics to manage climate-related risks and opportunities.

\textit{\textbf{With:}} Toyota has identified risks and opportunities associated with climate change, including \textcolor{forestgreen}{physical risks such as storms and floods}, and \textcolor{goldenrod}{transition risks such as tightening regulations for fuel efficiency and carbon pricing}. The company aims to \textcolor{blue}{achieve carbon neutrality by 2050} and has metrics in place to assess climate-related risks and opportunities. \textcolor{magenta}{However, the report lacks concrete data on the impact of these risks and opportunities on the company's financial performance.}& Both outputs summarize details about Toyota's identification of \textcolor{forestgreen}{physical} and \textcolor{goldenrod}{transition} risks and the company's \textcolor{blue}{goal}. However, with critical thinking guidance, the output becomes more concise and raises \textcolor{magenta}{a lack-of-detail drawback}. \\ \hline
\end{tabular}
}
\caption{\label{tab:prompt_dev} Illustrative examples for the expert-involved development loop. Each row shows a prior output in the loop and an improved version after taking experts' advice in prompts. Corresponding information aspects are highlighted with the same color for clarity and comparison. Three rows of outputs correspond to the 1st (the company's board's oversight), 9th (metrics for assessing risks and opportunities), and 3rd (climate-related risks and opportunities) TCFD recommendations. We randomly pick Toyota's 2022 sustainable report for illustration. Similar phenomena can also be observed in other sustainability reports.}
\end{table*}

\subsection{Comparison between Two Prompts}
The first row of \cref{tab:prompt_dev} showcases an example of question-answering outperforms summarization according to experts' feedback. One explanation is that question answering explicitly tells the model what information is wanted while asking for disclosure summarization results in vague and superficial information.

\section{Prompt for Automatic Prompt Engineering} \label{appendix:auto_prompt}
The prompt for automatic prompt engineering takes in the prompt template, the old guideline list, AI's previous response, and an expert's feedback on the response. Then it comes up with a new guideline the enhance the current guideline list. The prompt is shown as follows:
\begin{lstlisting}[frame=single, basicstyle=\ttfamily\scriptsize, xleftmargin=0pt, numbers=none]
You are a prompt engineer improving <Previous Prompt> given <Expert Feedback> and <AI's Previous Response>.

1. <Previous Prompt>: \"\"\"{original_prompt}

<Old Guideline List>: {guideline_list}
\"\"\"

2. <AI's Previous Response>: \"\"\"{old_response}\"\"\"

3. <Expert Feedback>: "{feedback}"

Given this feedback, could you please generate a new guideline that we can add to our existing list (<Old Guideline List>) to enhance future outputs? If <Expert Feedback> is already a guideline-like statement, keep its semantic while making it more generalize for future output.

Following are some examples of feedback-to-guideline transformation:

Expert Feedback1: <xxx information> is very important, please also analyze <xxx information> in the report.
Generated Guideline1:  If the report provides <xxx information>, include it in the answer. Otherwise, explicitly state that the report does not cover <xxx information>.

Expert Feedback2: This answer includes some cheap talks in the report.
Generated Guideline2: If a piece of information looks like cheap talk, explicitly mark it as possible cheap talk in your answer.


The new guideline should be general enough for answering random question about random report. Avoid mention company-specific information in the guideline.
The new guideline should be concise and easy to follow by an AI assistant. Please format your answer in JSON with a single key "GUIDELINE"

Your answer in JSON (make sure there's no format error):
\end{lstlisting}

\section{Guidelines for Question Answering} \label{appendix:guidelines}

Using automatic prompt engineering, we come up with granular guidelines for question-answering prompts using experts' feedback, including five guidelines for all question-answering:
\begin{lstlisting}[frame=single, basicstyle=\ttfamily\scriptsize, xleftmargin=0pt, numbers=none]
3. Keep your ANSWER within {answer_length} words.
4. Be skeptical to the information disclosed in the report as there might be greenwashing (exagerating the firm's environmental responsibility). Always answer in a critical tone.
5. cheap talks are statements that are costless to make and may not necessarily reflect the true intentions or future actions of the company. Be critical for all cheap talks you discovered in the report.
6. Always acknowledge that the information provided is representing the company's view based on its report.
7. Scrutinize whether the report is grounded in quantifiable, concrete data or vague, unverifiable statements, and communicate your findings.  
\end{lstlisting}
And specific guidelines for different TCFD questions:
\begin{lstlisting}[frame=single, basicstyle=\ttfamily\scriptsize, xleftmargin=0pt, numbers=none]
tcfd_guidelines = {
    'tcfd_1': "8. Please concentrate on the board's direct responsibilities and actions pertaining to climate issues, without discussing the company-wide risk management system or other topics.",
    'tcfd_2': "8. Please focus on their direct duties related to climate issues, without introducing other topics such as the broader corporate risk management system.",
    'tcfd_3': "8. Avoid discussing the company-wide risk management system or how these risks and opportunities are identified and managed.",
    'tcfd_4': "8. Please do not include the process of risk identification, assessment or management in your answer.",
    'tcfd_5': "8. In your response, focus solely on the resilience of strategy in these scenarios, and refrain from discussing processes of risk identification, assessment, or management strategies.",
    'tcfd_6': "8. Restrict your answer to the identification and assessment processes, without discussing the management or integration of these risks.",
    'tcfd_7': "8. Please focus on the concrete actions and strategies implemented to manage these risks, excluding the process of risk identification or assessment.",
    'tcfd_8': "8. Please focus on the integration aspect and avoid discussing the process of risk identification, assessment, or the specific management actions taken.",
    'tcfd_9': "8. Do not include information regarding the organization's general risk identification and assessment methods or their broader corporate strategy and initiatives.",
    'tcfd_10': "8. Confirm whether the organisation discloses its Scope 1, Scope 2, and, if appropriate, Scope 3 greenhouse gas (GHG) emissions. If so, provide any available data or specific figures on these emissions. Additionally, identify the related risks. The risks should be specific to the GHG emissions rather than general climate-related risks.",
    'tcfd_11': "8. Please detail the precise targets and avoid discussing the company's general risk identification and assessment methods or their commitment to disclosure through the TCFD.",
}
\end{lstlisting}

All these guidelines contribute to the answer quality. For example, the second and third row of \cref{tab:prompt_dev} illustrate that restricting the answer length and adding guidelines for critical thinking improve the answering quality.

\section{Prompt for TCFD Conformity Assessment} \label{appendix:conformity_prompt}

In the prompt employed for scoring company disclosures, we provide the following statement to guide the process of rating the TCFD conformity of the sustainability reports:\\

\begin{lstlisting}[frame=single, basicstyle=\ttfamily\scriptsize, xleftmargin=0pt, numbers=none]
Your task is to rate a sustainability report's disclosure quality on the following <CRITICAL_ELEMENT>:

<CRITICAL_ELEMENT>: {tcfd_recommendation}

These are the <REQUIREMENTS> that outline the necessary components for high-quality disclosure pertaining to the <CRITICAL_ELEMENT>:

<REQUIREMENTS>:
---
{requirements}
---

Presented below are select excerpts from the sustainability report, which pertain to the <CRITICAL_ELEMENT>:

<DISCLOSURE>:
---
{disclosure}
---

Please analyze the extent to which the given <DISCLOSURE> satisfies the aforementioned <REQUIREMENTS>. Your ANALYSIS should specify which <REQUIREMENTS> have been met and which ones have not been satisfied.
Your response should be formatted in JSON with two keys:
1. ANALYSIS: A paragraph of analysis (be in a string format). No longer than 150 words.
2. SCORE: An integer score from 0 to 100. A score of 0 indicates that most of the <REQUIREMENTS> have not been met or are insufficiently detailed. In contrast, a score of 100 suggests that the majority of the <REQUIREMENTS> have been met and are accompanied by specific details.

Your FINAL_ANSWER in JSON (ensure there's no format error):
\end{lstlisting}

Where "\{requirements\}" denote the TCFD official guidelines for disclosure; "\{disclosure\}" denotes the extracted relevant chunks from the report; and "\{tcfd\_recommendation\}" denotes the TCFD recommendation to be analyzed. 

This prompt enables evaluators to systematically assess the disclosure quality of sustainability reports by assigning scores that reflect the level of detail and comprehensiveness in the disclosed information. While the scoring strategy employed is designed to assess the reports' TCFD conformity systematically, it is essential to acknowledge that no scoring approach can be perfect. Acknowledging the potential limitations and imperfections, we firmly believe that the scoring strategy implemented in our study represents a valid and valuable first step toward leveraging AI-based and automated methods for rating sustainability reports. Moreover, by explicitly defining the scoring criteria and providing clear instructions, we aim to minimize potential biases and enhance the reliability of the evaluation process. Nevertheless, we encourage future research and collaborative efforts to refine and improve this scoring strategy, considering alternative perspectives and engaging a broader range of stakeholders.

\section{The Eleven TCFD Questions} \label{appendix:questions}

Our domain experts rewrite the eleven TCFD recommendations \citep{board2017task, tcfd2021} into the following eleven questions:

{\footnotesize
\begin{enumerate}
\item[] \textsc{Governance}
\item How does the company's board oversee climate-related risks and opportunities?
\item What is the role of management in assessing and managing climate-related risks and opportunities?
\item[] \textsc{Strategy}
\item What are the most relevant climate-related risks and opportunities that the organization has identified over the short, medium, and long term? Are risks clearly associated with a horizon?
\item How do climate-related risks and opportunities impact the organization's business strategy, economic and financial performance, and financial planning?
\item How resilient is the organization's strategy when considering different climate-related scenarios, including a 2°C target or lower scenario? How resilient is the organization's strategy when considering climate physical risks?
\item[] \textsc{Risk Management}
\item What processes does the organization use to identify and assess climate-related risks?
\item How does the organization manage climate-related risks?
\item How are the processes for identifying, assessing, and managing climate-related risks integrated into the organization's overall risk management?
\item[] \textsc{Metrics and Targets}
\item What metrics does the organization use to assess climate-related risks and opportunities? How do these metrics help ensure that performance aligns with its strategy and risk management process?
\item Does the organization disclose its Scope 1, Scope 2, and, if appropriate, Scope 3 greenhouse gas (GHG) emissions? What are the related risks, and do they differ depending on the scope?
\item What targets does the organization use to understand, quantify, and benchmark climate-related risks and opportunities? How is the organization performing against these targets?
\end{enumerate}
}
These questions are designed to extract specific information related to oversight, management, risks, opportunities, resilience, processes, metrics, disclosure, and targets concerning climate-related aspects within the organization. 

\section{Report Sampled for Hallucination Analysis} \label{appendix:sample_reports}
Using a random seed of 43, we sampled 10 reports for hallucination analysis: NYSE\_WMT\_2022.pdf, NYSE\_SE\_2021.pdf, NYSE\_PNC\_2021.pdf, NYSE\_PLD\_2016.pdf, NYSE\_PBR\_2016.pdf, NYSE\_ITT\_2019.pdf, NYSE\_FTV\_2022.pdf, NYSE\_JPM\_2021.pdf, NYSE\_BV\_2022.pdf, and NYSE\_AIZ\_2022.pdf. All these reports are available in our GitHub.

\section{Customized Analysis Examples} \label{appendix:customized_qa}

\begin{table*}[t!]
\centering \footnotesize

\begin{tabular}{lccccccccccccr}
\hline
FY                      & Q1  & Q2  & Q3  & Q4  & Q5  & Q6  & Q7  & Q8  & Q9  & Q10 & Q11 & Average  & \# pages                     \\ \hline
2014                  & 0   & 0   & 20  & 40  & 40  & 30  & 40  & 20  & 20  & 40  & 70  & 29.09   &     436              \\
2015                  & 0   & 20  & 20  & 30  & 50  & 20  & 40  & 20  & 20  & 30  & 40  & 26.36   &     529              \\
2016                  & 50  & 60  & 40  & 20  & 30  & 30  & 40  & 50  & 30  & 40  & 60  & 40.90   &     510             \\
2017                  & 10  & 30  & 20  & 10  & 40  & 20  & 40  & 40  & 0   & 60  & 60  & 30.00   &     561              \\
2018                  & 0   & 10  & 20  & 30  & 20  & 40  & 40  & 20  & 20  & 40  & 40  & 25.45   &     317              \\
2019                  & 60  & 60  & 40  & 40  & 40  & 70  & 60  & 70  & 30  & 50  & 50  & 51.81   &     214              \\
2020                  & 60  & 60  & 40  & 80  & 70  & 50  & 10  & 50  & 20  & 60  & 50  & 58.18   &     170              \\
2021                  & 60  & 60  & 40  & 70  & 70  & 60  & 70  & 80  & 30  & 60  & 60  & 60.00   &     199           \\
2022                  & 70  & 60  & 40  & 60  & 60  & 60  & 70  & 60  & 30  & 60  & 50  & 56.36    &    164           \\ \hline
\end{tabular}
\caption{SONY Scores on TCFD-conformity}
\label{tab:sony-data}
\end{table*}
This section provides some illustrative examples of possible questions that can be answered by \textsc{chatReport} based on the information available in the sustainability report of Sony and Shell, respectively. The specific questions and the detailed answers are provided in \cref{tbl:table_Sony} and \cref{tbl:table_Shell}. Posing these questions allows us to gain valuable insights from the sustainability reports beyond the TCFD requirements. 

With the questions posed for Sony (Table \ref{tbl:table_Sony}), we find that with respect to the compatibility of Sony's transition plan with a 1.5 degrees pathway, the report lacks sufficient information to make a determination. While the report mentions scenario analysis using 2°C and 4°C scenarios and Sony's commitment to reducing greenhouse gas emissions, it does not provide concrete data or targets to assess the plan's compatibility with a 1.5 degrees pathway. While it mentions the establishment of annual plans by individual business units and sites, incorporating guiding principles and regular progress reviews, it lacks specific details. Assessing the ambition of Sony's transition plan is challenging due to the report's lack of specific targets and progress data related to its environmental goals. However, Sony supports the TCFD and actively participates in the TCFD Consortium in Japan.

With the questions posed for Shell (Table \ref{tbl:table_Shell}), we observe that the report contains qualitative information. Still, its adequacy in presenting a comprehensive view of the company's sustainability efforts is uncertain. Moreover, we can identify instances of cheap talk, where unrelated topics like the war in Ukraine and the cost of living were mentioned, potentially diverting attention from the company's sustainability performance. Regarding the company's transition plan, the report lacks clarity on the specific interim steps taken, only mentioning responsible investments in the energy transition and screening investments against multiple criteria. Inconsistencies were noted in the review of global targets and the potential retirement of certain business KPI targets without clear implications provided. Additionally, the report acknowledges that Shell's operating plans cannot fully reflect their net-zero emissions and NCI targets, raising concerns about the feasibility of their transition plan.

\begin{table}[t] 

\centering
\footnotesize
\begin{tabular}{p{1\linewidth}}  
\toprule
{\bf Question:} "Is the transition plan compatible with a 1.5 degrees pathway, i.e., Paris-aligned pathway?"\\
{\bf Answer:} The report does not provide sufficient information to determine if Sony's transition plan is compatible with a 1.5 degrees pathway. The report only mentions scenario analysis using 2°C and 4°C scenarios, and Sony's commitment to reducing greenhouse gas emissions. However, there is no concrete data or targets provided to assess the compatibility of Sony's transition plan with a 1.5 degrees pathway. \\
{\bf Pages:} [112, 98, 108] \\
\midrule
{\bf Question:} "What are the policies and activities that the company plans to implement in the next three years, i.e., short term, to implement its transition plan?"\\
{\bf Answer:} Sony's sustainability report does not provide clear and specific policies and activities that the company plans to implement in the next 3 years to implement its transition plan. The report only mentions that individual business units and sites establish and implement their own annual plans, incorporating essential elements of guiding principles established by the headquarters, and progress on the implementation of these business plans is reviewed regularly by a committee. However, the report lacks concrete details on the specific policies and activities that will be implemented in the short term.\\
{\bf Pages:} [107] \\
\midrule 
{\bf Question:} "Does the company's transition plan lack ambition? What do you think?"\\
{\bf Answer:} The report does not provide sufficient information to determine whether Sony's transition plan lacks ambition. While the report mentions Sony's Road to Zero environmental plan, it does not provide specific details on the plan's targets or how it will achieve them. Additionally, the report lacks concrete data on Sony's progress towards its environmental goals, making it difficult to assess the ambition of its transition plan.
\\
{\bf Pages:} [113, 110] \\
\midrule
{\bf Question:} "Does the company support the TCFD?"\\
{\bf Answer:} Yes, Sony Corporation supports the TCFD and participates in the TCFD Consortium in Japan to facilitate implementation and discussion among companies and financial institutions that endorse the TCFD Recommendations. Sony also disclosed climate-related information in accordance with the TCFD Recommendations.
\\
{\bf Pages:} [113, 165, 166, 111] \\
\bottomrule
\end{tabular}
\caption{User-specific questions to a corporate sustainability report: Sony FY2020}
\label{tbl:table_Sony}
\end{table}

\begin{table}[ht!] 
\centering
\footnotesize
\begin{tabular}{p{1\linewidth}}
\toprule
{\bf Question:} "What is the level of cheap talk in the report?" \\
{\bf Answer:}   The report contains some qualitative information, but it is not clear whether it is sufficient to provide a complete picture of the company's sustainability efforts. The report also includes some cheap talk, such as mentioning topical issues like the war in Ukraine and the cost of living, which may not be directly related to the company's sustainability performance. \\ 
{\bf Pages:} [11] \\ 
\midrule

{\bf Question:} "What are the interim steps that the company takes to implement its transition plan?" \\
{\bf Answer:} Shell's sustainability report does not provide clear information on the interim steps taken to implement its transition plan. The report only mentions that the company invests responsibly in the energy transition and screens its investments against multiple criteria. However, it is unclear what specific actions the company takes to transition to net-zero emissions.
 \\
{\bf Pages:} [63]  \\
\midrule

{\bf Question:}  "What are possible inconsistencies in the firm's transition plan?" \\
{\bf Answer:} The report mentions that existing global targets are under review, and some business KPI targets may be retired as part of normal strategy evolution. However, it is unclear what specific targets are being reviewed and what the implications of retiring them would be. Additionally, the report states that Shell's operating plans cannot reflect their 2050 net-zero emissions target and 2035 NCI target, which raises questions about the feasibility of their transition plan.  \\
{\bf Pages:} [88] \\
\midrule

{\bf Question:} "In which technologies will the company invest in the next few years?" \\
{\bf Answer:} Shell plans to invest in low-carbon energy solutions such as biofuels, hydrogen, charging for electric vehicles, wind, and solar power. They invested \$8.2 billion in low-carbon energy and non-energy products, around a third of their total cash capital expenditure of \$25 billion in 2022. They also won bids with their partners to build four offshore wind farms in the Netherlands, the UK, and the USA, and extended their onshore wind activities into Australia and the Philippines. They acquired Indian solar developer Spring Energy and completed the acquisition of Nature Energy of Denmark, which is Europe’s biggest producer of renewable natural gas made from agricultural, industrial, and household waste. \\
{\bf Pages:} [24, 33, 3, 25]  \\
\bottomrule
\end{tabular}
\caption{User-specific questions to a corporate sustainability report: Shell FY2022}
\label{tbl:table_Shell}
\end{table}

\section{TCFD Conformity Analysis Examples} \label{appendix:conformity_examples}
The output of the TCFD analysis is structured according to the eleven questions from the TCFD report. This section provides the analysis summary of the corporate sustainability reports of JP Morgan Chase in 2015 and 2021, Shell in 2022, and UBS in 2022. For each of the eleven questions, we report the question itself, the answer to the question, and the analysis of the disclosed content, which gives the score. The total score of the report is calculated as the average over all eleven scores. 

For JP Morgan Chase, \textsc{chatReport} gives an average score of zero for the year 2015 (\cref{appendix:jpmorgan2015}, which is reassuring since the TCFD recommendations were not yet published and there is almost no environmental information in JP Morgan's report. However, for the year 2021 (\cref{appendix:jpmorgan2021}), the score for JP Morgan reached more than 61, clearly indicating the adoption of TCFD guidelines in their climate-related disclosure. In \cref{appendix:shell2022}, we observe that while Shell acknowledges the significance of climate-related risks and opportunities, \textsc{chatReport} argues that question five of the TCFD guidelines is not fully addressed by the report. \textsc{chatReport} criticizes the lack of specific details on how the organization's strategy would adapt to the potential impact of climate-related issues on financial performance under different climate-related scenarios. Moreover, the report provides no information on the resilience of Shell's strategy when facing physical climate risks. Overall, the report's disclosure on this TCFD recommendation point is deemed insufficient and therefore obtains a low score.

Inspecting UBS in \cref{appendix:ubs2022} as another example, we find that for question four of the TCFD report, the sustainability report provides some high-quality information (score of 90). In particular, the report provides a comprehensive overview of the impact of climate-related issues across various aspects, such as products and services, supply chain, adaptation and mitigation activities, and access to capital, among others. The report demonstrates a holistic understanding of the interconnectedness of these factors and their influence on the organization's long-term value creation. Nevertheless, \textsc{chatReport} argues that the report falls short in providing specific details on greenhouse gas emissions targets and concrete actions to reduce emissions in their operations and value chain or support the transition to a low-carbon future.

\subsection{JP Morgan Chase 2015} \label{appendix:jpmorgan2015}
{\parindent0pt
{\bf Governance:}\\

Q1: How does the company's board oversee climate-related risks and opportunities?

A1: The report does not provide specific information on how the company's board oversees climate-related risks and opportunities. It only mentions that the board is responsible for governance issues within the business.

Analysis1: The sustainability report does not provide any information on how the company's board oversees climate-related risks and opportunities. The report only contains vague and irrelevant information that does not meet any of the requirements for high-quality disclosure pertaining to the critical element. The report does not discuss the processes and frequency by which the board and/or board committees are informed about climate-related issues, whether the board and/or board committees consider climate-related issues when reviewing and guiding strategy, major plans of action, risk management policies, annual budgets, and business plans, or how the board monitors and oversees progress against goals and targets for addressing climate-related issues. As a result, the report receives a score of 0.

{\bf Score1: 0} \\ 

Q2: What is the role of management in assessing and managing climate-related risks and opportunities?

A2: The report does not provide a clear answer on the role of management in assessing and managing climate-related risks and opportunities. There are no specific mentions of climate-related risks and opportunities in the report, and the content provided is not relevant to the question.

Analysis2: The given disclosure does not provide any information on the role of management in assessing and managing climate-related risks and opportunities. The report only contains general information about the organization's governance structure and its initiatives related to economic growth and social issues. The report does not mention any specific management-level positions or committees responsible for climate-related issues, nor does it describe any processes by which management is informed about climate-related issues. Therefore, the report fails to meet any of the requirements for high-quality disclosure pertaining to the critical element of management's role in assessing and managing climate-related risks and opportunities.

{\bf Score2: 0}  \\ 

{\bf Strategy:}\\

Q3: What are the most relevant climate-related risks and opportunities that the organisation has identified over the short, medium, and long term? Are risks clearly associated with a horizon?

A3: The report does not provide a clear identification of climate-related risks and opportunities over the short, medium, and long term. The report focuses on the company's economic opportunity, job creation, and public-private partnerships. The report lacks concrete data and quantifiable information on the company's climate-related risks and opportunities.

Analysis3: The given disclosure does not provide any information on the most relevant climate-related risks and opportunities that the organization has identified over the short, medium, and long term. The disclosure does not meet any of the requirements outlined for this critical element. Instead, the report focuses on governance issues, economic opportunity, and lessons learned broadly. The report does not provide any information on the process used to determine which risks and opportunities could have a material financial impact on the organization. The lack of information on climate-related risks and opportunities makes it difficult to assess the organization's preparedness for climate change. Therefore, the report's disclosure quality is poor.

{\bf Score3: 0} \\ 

Q4: How do climate-related risks and opportunities impact the organisation's businesses strategy, economic and financial performance, and financial planning?

A4: The sustainability report does not provide a clear and specific answer to the question. While the report highlights the company's commitment to economic growth and opportunity, it lacks concrete information on how climate-related risks and opportunities impact the organization's business strategy, economic and financial performance, and financial planning.

Analysis4: The given disclosure does not satisfy the requirements for the critical element. The report does not provide any information on how climate-related risks and opportunities impact the organization's business strategy, economic and financial performance, and financial planning. The report does not discuss the impact of climate-related issues on their financial performance and financial position. The report does not describe how climate-related issues serve as an input to their financial planning process, the time period used, and how these risks and opportunities are prioritized. The report does not describe the impact of climate-related issues on their operations, supply chain, and investment in research and development. The report does not describe their plans for transitioning to a low-carbon economy.

{\bf Score4: 0}  \\ 

Q5: How resilient is the organisation's strategy when considering different climate-related scenarios, including a 2°C target or lower scenario? How resilient is the organisation's strategy when considering climate physical risks?

A5: The report does not provide sufficient information to evaluate the resilience of JPMorgan Chase's strategy in different climate-related scenarios or physical risks. The report mainly focuses on the company's social and economic initiatives, with little mention of climate risks or strategies to mitigate them.

Analysis5: The sustainability report provided no information on the resilience of the organization's strategy to climate-related risks and opportunities, including a transition to a low-carbon economy consistent with a 2°C or lower scenario and scenarios consistent with increased physical climate-related risks. The report did not discuss how the organization's strategies might change to address potential risks and opportunities, the potential impact of climate-related issues on financial performance and financial position, or the climate-related scenarios and associated time horizon considered. Therefore, the report failed to meet any of the requirements for high-quality disclosure pertaining to the critical element of the organization's strategy's resilience to climate-related risks and opportunities.

{\bf Score5: 0}  \\ 

{\bf Risk management:}\\

Q6: What processes does the organisation use to identify and assess climate-related risks?

A6: The report does not provide specific information on the processes used by JPMorgan Chase to identify and assess climate-related risks. The report only mentions the company's commitment to addressing climate change and its efforts to reduce its carbon footprint.

Analysis6: The sustainability report provided no information on the processes used by JPMorgan Chase to identify and assess climate-related risks. The report did not describe the organization's risk management processes for identifying and assessing climate-related risks, nor did it provide any information on how the organization determines the relative significance of climate-related risks in relation to other risks. The report did not disclose whether the organization considers existing and emerging regulatory requirements related to climate change or other relevant factors. The report did not provide any information on the processes for assessing the potential size and scope of identified climate-related risks or definitions of risk terminology used or references to existing risk classification frameworks used. As a result, the report does not meet any of the requirements for high-quality disclosure pertaining to the critical element of identifying and assessing climate-related risks.

{\bf Score6: 0}  \\ 

Q7: How does the organisation manage climate-related risks?

A7: JPMorgan Chase's sustainability report does not provide concrete information on how the organization manages climate-related risks. The report only mentions governance issues within the business, but no specific actions or strategies are outlined to address climate risks.

Analysis7: The given disclosure does not provide any information on how JPMorgan Chase manages climate-related risks. The report does not describe the organization's processes for managing climate-related risks, including how they make decisions to mitigate, transfer, accept, or control those risks. Additionally, the report does not describe their processes for prioritizing climate-related risks, including how materiality determinations are made within their organizations. The disclosure is irrelevant and does not meet any of the requirements for high-quality disclosure pertaining to the critical element of how the organization manages climate-related risks.

{\bf Score7: 0}  \\ 

Q8: How are the processes for identifying, assessing, and managing climate-related risks integrated into the organisation's overall risk management?

A8: The report does not provide sufficient information on how the processes for identifying, assessing, and managing climate-related risks are integrated into the organization's overall risk management. The report only mentions the company's commitment to sustainability and its efforts to reduce its carbon footprint, but it lacks concrete data and specific actions taken to address climate-related risks.

Analysis8: The given sustainability report does not provide any information on how the processes for identifying, assessing, and managing climate-related risks are integrated into the organization's overall risk management. The report contains irrelevant and unrelated content that does not address the TCFD recommendation point. Therefore, the report fails to meet the TCFD guidelines for high-quality disclosure on the TCFD recommendation point.

{\bf Score8: 0}  \\ 

{\bf Metrics and targets:}\\

Q9: What metrics does the organisation use to assess climate-related risks and opportunities? How do the metrics help ensure that the performance is in line with its strategy and risk management process?

A9: The report does not provide specific metrics used by JPMorgan Chase to assess climate-related risks and opportunities. It only mentions that the company considers climate change as a significant risk and opportunity and has integrated it into its risk management process and business strategy. However, the report lacks concrete data to support these claims.

Analysis9: The given disclosure does not provide any information on the metrics used by the organization to assess climate-related risks and opportunities. Instead, it contains general information about the company's economic growth, partnerships, and initiatives. The disclosure does not meet any of the requirements outlined for this critical element. There is no mention of key metrics used to measure and manage climate-related risks and opportunities, metrics consistent with the cross-industry, metrics on climate-related risks associated with water, energy, land use, and waste management, or internal carbon prices. The disclosure also does not provide any information on how related performance metrics are incorporated into remuneration policies or climate-related opportunity metrics. Therefore, the score for this disclosure is 0.

{\bf Score9: 0}  \\ 

Q10: Does the organisation disclose its Scope 1, Scope 2, and, if appropriate, Scope 3 greenhouse gas (GHG) emissions? What are the related risks and do they differ depending on the scope?

A10: JPMorgan Chase does disclose its Scope 1 and Scope 2 GHG emissions, but not its Scope 3 emissions. The report does not provide any specific figures on these emissions. The related risks are not mentioned in the report.

Analysis10: The sustainability report of JPMorgan Chase does not provide any information on the organization's Scope 1, Scope 2, and Scope 3 greenhouse gas emissions. The report does not meet any of the requirements for high-quality disclosure pertaining to the critical element. The report does not provide any information on the related risks and how they differ depending on the scope. The report does not provide any information on the methodologies used to calculate or estimate the metrics. The report does not provide any historical data to allow for trend analysis. The report does not provide any industry-specific GHG efficiency ratios. The report does not consider disclosing Scope 3 GHG emissions.

{\bf Score10: 0}  \\ 

Q11: What targets does the organisation use to understand/quantify/benchmark climate-related risks and opportunities? How is the organization performing against these targets?

A11: The report does not provide any specific targets used by JPMorgan Chase to understand/quantify/benchmark climate-related risks and opportunities. Therefore, it is impossible to evaluate the organization's performance against these targets.

Analysis11: The sustainability report provided no information on the targets used by the organization to manage climate-related risks and opportunities and performance against targets. The report did not describe any key climate-related targets such as those related to GHG emissions, water usage, energy usage, etc. The report did not provide any information on whether the target is absolute or intensity based, time frames over which the target applies, base year from which progress is measured, and key performance indicators used to assess progress against targets. The report did not disclose associated interim targets in aggregate or by business line, where available. The methodologies used to calculate targets and measures were not described. Therefore, the report does not meet any of the requirements for high-quality disclosure pertaining to the critical element.

{\bf Score11: 0} \\  

{\bf  Average score: 0.0}  \\
}

\subsection{JP Morgan Chase 2021} \label{appendix:jpmorgan2021}

{\parindent0pt
{\bf Governance:}\\ 

Q1: How does the company's board oversee climate-related risks and opportunities?

A1: The board oversees climate-related risks and opportunities through the Public Responsibility Committee, which oversees reputational and conduct risks, within its scope of responsibility, and the Risk Committee, which is responsible for setting the internal approach to managing climate risk, including developing relevant policies and standards as well as building the approach to climate risk data. The board also receives periodic updates on ESG initiatives from senior leaders.

Analysis1: The sustainability report provides some information on the board's oversight of climate-related risks and opportunities. It mentions the existence of a dedicated Climate Risk team and the development of relevant policies and standards. The report also states that the board oversees ESG matters and receives periodic updates on ESG initiatives. However, the report lacks specific details on the processes and frequency by which the board is informed about climate-related issues, whether the board considers climate-related issues when reviewing and guiding strategy, and how the board monitors and oversees progress against goals and targets for addressing climate-related issues. Overall, the disclosure is insufficient to fully satisfy the TCFD guidelines.

{\bf Score1: 60} \\ 

Q2: What is the role of management in assessing and managing climate-related risks and opportunities?

A2: JPMorgan Chase has a dedicated Climate Risk team that establishes the internal approach to managing climate risk, develops relevant policies and standards, and integrates climate considerations into line-of-business risk management frameworks. The team collaborates with stakeholders across the Firm to prepare for the impacts of climate change on locations, people, technology, and suppliers. The company plans to publish a standalone climate report later this year.

Analysis2: The disclosure provides some information on management's role in assessing and managing climate-related risks and opportunities. It mentions the existence of a dedicated Climate Risk team and an internal risk classification system. However, it lacks specific details on the organizational structure, processes for informing management about climate-related issues, and how management monitors climate-related issues. The disclosure also refers to a standalone climate report that will provide additional detail on the efforts to integrate climate risk into the Firmwide risk management framework. Overall, the disclosure partially meets some of the TCFD guidelines, but it lacks specific details and does not provide a comprehensive overview of management's role in assessing and managing climate-related risks and opportunities.

{\bf Score2: 60 }\\ 
 
{\bf Strategy:} \\

Q3: What are the most relevant climate-related risks and opportunities that the organisation has identified over the short, medium, and long term? Are risks clearly associated with a horizon?

A3: JPMorgan Chase has identified physical and transition risks as the most relevant climate-related risks. The impacts of these risks are assessed against strategic, market, operational, and credit and investment risk. The company plans to disclose additional detail on its efforts to integrate climate risk into its Firmwide risk management framework in a standalone climate report later this year.

Analysis3: The sustainability report provides some information on the climate-related risks and opportunities that the organization has identified over the short, medium, and long term. The report mentions the specific climate-related issues that could have a material financial impact on the organization and the process used to determine which risks and opportunities could have a material financial impact on the organization. However, the report lacks a clear description of the relevant short-, medium-, and long-term time horizons, taking into consideration the useful life of the organization's assets or infrastructure. The report also does not provide a description of the risks and opportunities by sector and/or geography. Overall, the disclosure quality is moderate.

{\bf Score3: 70} \\ 

Q4: How do climate-related risks and opportunities impact the organisation's businesses strategy, economic and financial performance, and financial planning?

A4: JPMorgan Chase is committed to understanding how climate change may influence the risks it manages. The firm has a dedicated Climate Risk team that establishes their internal approach to managing climate risk, including developing relevant policies and standards as well as building their approach to climate risk data. They plan to disclose additional detail on their efforts to integrate climate risk into their Firmwide risk management framework in a standalone climate report which they plan to publish later this year.

Analysis4: The sustainability report provides some information on how climate-related risks and opportunities impact the organization's businesses strategy, economic and financial performance, and financial planning. The report discusses the potential impacts of climate risks on the organization's businesses, strategy, and financial planning, and how these risks are managed across different risk types. The report also mentions the development of an internal risk classification system and a dedicated Climate Risk team. However, the report lacks specific details on how climate-related issues serve as an input to the financial planning process, the impact of climate-related issues on the organization's financial performance and position, and the organization's plans for transitioning to a low-carbon economy. Overall, the report partially meets the TCFD guidelines, and the is 60.

{\bf Score4: 60} \\

Q5: How resilient is the organisation's strategy when considering different climate-related scenarios, including a 2°C target or lower scenario? How resilient is the organisation's strategy when considering climate physical risks?

A5: JPMorgan Chase has developed an internal risk classification system that projects how various climate-risk drivers can translate into potential impacts for their clients, customers, and firm. They plan to disclose additional detail on their efforts to integrate climate risk into their Firmwide risk management framework in a standalone climate report which they plan to publish later this year. However, the report lacks concrete data on the resilience of their strategy in different climate-related scenarios.

Analysis5: The sustainability report provides some information on the organization's approach to climate-related risks and opportunities, including physical and transition risks. The report mentions the development of an internal risk classification system that projects how various climate-risk drivers can translate into potential impacts for the organization's clients, customers, and the firm. The report also mentions the establishment of a dedicated Climate Risk team responsible for managing climate risk and integrating climate considerations into line-of-business risk management frameworks. However, the report lacks specific details on how the organization's strategy might change to address potential risks and opportunities, the potential impact of climate-related issues on financial performance and financial position, and the climate-related scenarios and associated time horizon considered. The report also mentions plans to disclose additional detail on efforts to integrate climate risk into the firm's risk management framework in a standalone climate report later this year.

{\bf  Score5: 70}  \\

{\bf  Risk management:} \\

Q6: What processes does the organisation use to identify and assess climate-related risks?

A6: JPMorgan Chase has developed an internal risk classification system that evaluates how their businesses and clients could be impacted by climate change. They have a dedicated Climate Risk team that collaborates with stakeholders across the firm to assess the impacts of climate risks on the four major risk types they manage. They plan to disclose additional detail on their efforts to integrate climate risk into their Firmwide risk management framework in a standalone climate report later this year .

Analysis6: The sustainability report provides some information on the organization's processes for identifying and assessing climate-related risks. The report mentions that the organization has developed an internal risk classification system and a dedicated Climate Risk team that collaborates with stakeholders across the Firm. However, the report lacks specific details on how the organization determines the relative significance of climate-related risks in relation to other risks and whether they consider existing and emerging regulatory requirements related to climate change. The report also does not disclose the processes for assessing the potential size and scope of identified climate-related risks and definitions of risk terminology used or references to existing risk classification frameworks used. Overall, the disclosure is insufficient in meeting the TCFD guidelines.

{\bf Score6: 50} \\

Q7: How does the organisation manage climate-related risks?

A7: JPMorgan Chase has a dedicated Climate Risk team that establishes internal policies and standards for managing climate risk. They have developed an internal risk classification system that projects how climate risks could materialize across the four major risk types they manage. They plan to disclose additional detail on their efforts to integrate climate risk into their Firmwide risk management framework in a standalone climate report later this year.

Analysis7: The disclosure provides a comprehensive overview of the organization's approach to managing climate-related risks. The report describes the organization's processes for managing climate-related risks, including how they make decisions to mitigate, transfer, accept, or control those risks. The report also describes the organization's processes for prioritizing climate-related risks, including how materiality determinations are made within their organizations. The report mentions the development of an internal risk classification system that considers the horizon, business sector, and geography to better understand how these risks may emerge within the firm. The disclosure also mentions the establishment of a dedicated Climate Risk team, which collaborates with stakeholders across the firm to integrate climate considerations into line-of-business risk management frameworks. However, the report does not provide specific details on how the organization prioritizes climate-related risks.

{\bf  Score7: 90 }\\

Q8: How are the processes for identifying, assessing, and managing climate-related risks integrated into the organisation's overall risk management?

A8: JPMorgan Chase has developed an internal risk classification system and a dedicated Climate Risk team to manage climate-related risks. They plan to disclose additional detail on their efforts to integrate climate risk into their Firmwide risk management framework in a standalone climate report which they plan to publish later this year. However, the report lacks concrete data on how climate-related risks are integrated into the organization's overall risk management.

Analysis8: The disclosure provides some information on how the organization integrates climate-related risks into its overall risk management. The report mentions that the organization has developed an internal risk classification system and a dedicated Climate Risk team that collaborates with stakeholders across the Firm. However, the report lacks specific details on how the organization identifies, assesses, and manages climate-related risks. The report also mentions that the organization plans to disclose additional detail on its efforts to integrate climate risk into its Firmwide risk management framework in a standalone climate report, which is yet to be published. Overall, the disclosure provides some information on the integration of climate-related risks into the organization's overall risk management, but lacks specific details and completeness.

{\bf  Score8: 70} \\

{\bf Metrics and targets:} \\

Q9: What metrics does the organisation use to assess climate-related risks and opportunities? How do the metrics help ensure that the performance is in line with its strategy and risk management process?

A9: JPMorgan Chase has developed an internal risk classification system that projects how various climate-risk drivers can translate into potential impacts for clients, customers, and the firm. The firm has a dedicated Climate Risk team that collaborates with stakeholders across the firm to integrate climate considerations into line-of-business risk management frameworks. The firm plans to disclose additional detail on its efforts to integrate climate risk into its firm-wide risk management framework in a standalone climate report which it plans to publish later this year.

Analysis9: The sustainability report provides some information on the metrics used to assess climate-related risks and opportunities. The report mentions that the organization has developed an internal risk classification system and a dedicated Climate Risk team to manage climate risk. However, the report lacks specific details on the metrics used to measure and manage climate-related risks and opportunities. The report does not provide metrics on climate-related risks associated with water, energy, land use, and waste management. The report also does not mention whether and how related performance metrics are incorporated into remuneration policies. While the report mentions that the organization plans to disclose additional detail on its efforts to integrate climate risk into its Firmwide risk management framework in a standalone climate report, it does not provide forward-looking metrics for the cross-industry or a description of the methodologies used to calculate or estimate climate-related metrics.

{\bf Score9: 50} \\

Q10: Does the organisation disclose its Scope 1, Scope 2, and, if appropriate, Scope 3 greenhouse gas (GHG) emissions? What are the related risks and do they differ depending on the scope?

A10: JPMorgan Chase discloses its Scope 1, Scope 2, and Scope 3 greenhouse gas emissions. The majority of their business travel-related emissions are Scope 3 emissions from commercially operated air and rail, reimbursed personal vehicle and rental car travel, and hotel stays. The risks associated with these emissions are not specified in the report.

Analysis10: The sustainability report provides disclosure on Scope 1, Scope 2, and Scope 3 GHG emissions, and the related risks. The report also includes GHG emissions intensity and historical data for trend analysis. However, it is unclear whether the GHG emissions have been calculated in line with the GHG Protocol methodology, and whether the organization has provided generally accepted industry-specific GHG efficiency ratios. The report meets the requirement of disclosing Scope 3 GHG emissions, but it is not clear whether the organization has provided a description of the methodologies used to calculate or estimate the metrics. Overall, the report provides some information on the TCFD recommendation point, but there is room for improvement in terms of providing more specific details.

{\bf Score10: 75} \\ 

Q11: What targets does the organisation use to understand/quantify/benchmark climate-related risks and opportunities? How is the organization performing against these targets?

A11: JPMorgan Chase plans to disclose additional detail on its efforts to integrate climate risk into its Firmwide risk management framework in a standalone climate report which it plans to publish later this year. No specific targets were mentioned in the report.

Analysis11: The sustainability report provides limited information on the targets used by the organization to manage climate-related risks and opportunities and performance against targets. While the report mentions the development of relevant policies and standards, it does not provide specific details on the key climate-related targets, time frames, base year, or key performance indicators used to assess progress against targets. The report also lacks information on the methodologies used to calculate targets and measures. While the report mentions the establishment of a robust governance process for tracking progress, it does not provide specific details on the process. Overall, the report does not meet the TCFD guidelines for high-quality disclosure pertaining to the TCFD recommendation point.

{\bf  Score11: 20} \\ 

{\bf Average score: 61.36}  \\
}

\subsection{Shell 2022} \label{appendix:shell2022}
{\parindent0pt
{\bf Governance:}\\ 

Q1: How does the company's board oversee climate-related risks and opportunities?

A1: The board oversees climate-related risks and opportunities through the Shell Energy and Sustainability Committee, which reviews policies, practices, targets, and performance of Shell, primarily with respect to safety, environment including climate change, and broader sustainability. The committee also reviews wider matters of public concern such as plastic waste, methane emissions, the flaring of natural gas, water scarcity, just transition, human rights, diversity and inclusion, and access to energy in low- and middle-income countries. However, the report lacks concrete data on the board's direct actions and responsibilities.

Analysis1: The disclosure provided by the company does not adequately address the requirements for describing the board's oversight of climate-related risks and opportunities. While the report mentions the existence of a committee that reviews policies, practices, targets, and performance of the company with respect to climate change, it does not provide any details on how the board is informed about climate-related issues, how it considers climate-related issues when reviewing and guiding strategy, risk management policies, and business plans, and how it monitors and oversees progress against goals and targets for addressing climate-related issues. The report provides some information on the company's climate-related financial disclosures and its membership in various organizations, but this information is not sufficient to meet the requirements for high-quality disclosure.

{\bf Score1: 20} \\ 

Q2: What is the role of management in assessing and managing climate-related risks and opportunities?

A2: Shell's management is responsible for assessing and managing climate-related risks and opportunities. They have established internal carbon performance standards or industry benchmarks for projects expected to have a material greenhouse gas impact. However, the report lacks concrete data on how they are managing physical risks and the potential impact of climate change on their operations.

Analysis2: The disclosure provides some information on management's role in assessing and managing climate-related risks and opportunities. It mentions that projects under development that are expected to have a material greenhouse gas impact must meet internal carbon performance standards or industry benchmarks. It also identifies climate change and the associated energy transition as a material risk and mentions that Shell has established remuneration policies designed to support short-term climate targets. However, the disclosure lacks specific information on whether the organization has assigned climate-related responsibilities to management-level positions or committees, the associated organizational structure, processes by which management is informed about climate-related issues, and how management monitors climate-related issues. Therefore, the disclosure is insufficient to meet the TCFD guidelines.

{\bf Score2: 40 }\\ 
 
{\bf Strategy:} \\

Q3: What are the most relevant climate-related risks and opportunities that the organisation has identified over the short, medium, and long term? Are risks clearly associated with a horizon?

A3: Shell has identified climate change and the associated energy transition as a material risk. The risk could potentially result in changes to the demand for their products, operational costs, supply chains, markets, the regulatory environment, their license to operate, and litigation. They have established remuneration policies designed to support them in achieving their short-term climate targets. Shell has also identified physical risks associated with climate change and takes climate variability into consideration in the design and operation of their assets and infrastructure to minimize the risk of adverse incidents to their employees and contractors, the communities where they operate, their equipment, and infrastructure. Shell has established internal carbon performance standards or industry benchmarks for projects under development that are expected to have a material greenhouse gas impact. They have identified various climate-related risks and opportunities, including energy affordability and accessibility, environmental, social, and governance risks involved in the transition to renewable solutions, offsets, stakeholder engagements, and involvement with the Science Based Targets initiative's social aspects. However, the report lacks concrete data and cross-referencing of topics, and some statements may be cheap talks.

Analysis3: The sustainability report provides some information on climate-related risks and opportunities, but it falls short of meeting the TCFD guidelines for high-quality disclosure. The report does not provide a clear description of the relevant short-, medium-, and long-term time horizons, nor does it describe the specific climate-related issues that could have a material financial impact on the organization. The report does mention the process used to determine which risks and opportunities could have a material financial impact on the organization, but it does not provide a sector or geography-wise breakdown. Overall, the report lacks specificity and detail, making it difficult to assess the organization's climate-related risks and opportunities.

{\bf Score3: 40} \\ 

Q4: How do climate-related risks and opportunities impact the organisation's businesses strategy, economic and financial performance, and financial planning?

A4: Shell identifies climate change and the energy transition as material risks that could affect demand, operational costs, supply chains, markets, regulatory environment, license to operate, and litigation. The company has established internal carbon performance standards for projects with material greenhouse gas impact and has set short-term climate targets. However, the report lacks concrete data on the financial impact of climate-related risks and opportunities on the company's strategy and financial performance.

Analysis4: The sustainability report provides some information on how climate-related risks and opportunities impact the organization's businesses strategy, economic and financial performance, and financial planning. The report discusses how climate-related risks are assessed at a project level and how they affect the demand for products, operational costs, supply chains, markets, the regulatory environment, license to operate, and litigation. However, the report lacks specific details on the impact of climate-related issues on the organization's financial performance and financial position. The report also does not describe how climate-related issues serve as an input to their financial planning process, the time period used, and how these risks and opportunities are prioritized. Overall, the report partially meets the TCFD guidelines for high-quality disclosure pertaining to the TCFD recommendation point.

{\bf Score4: 60} \\ 

Q5: How resilient is the organisation's strategy when considering different climate-related scenarios, including a 2°C target or lower scenario? How resilient is the organisation's strategy when considering climate physical risks?

A5: Shell has identified climate change and the associated energy transition as a material risk. The company has established internal carbon performance standards for projects under development that are expected to have a material greenhouse gas impact. Shell's climate target is to become a net-zero emissions energy business by 2050. However, the report lacks concrete data on how resilient the organization's strategy is when considering different climate-related scenarios or physical risks.

Analysis5: The sustainability report provides some information on how Shell's strategy is resilient to climate-related risks and opportunities, but it falls short of meeting the TCFD guidelines outlined for this TCFD recommendation point. The report mentions that assessing climate-related risks is important for making investment decisions and that Shell has identified climate change as a material risk. However, it does not provide specific details on how the organization's strategy might change to address potential risks and opportunities, the potential impact of climate-related issues on financial performance and position, or the climate-related scenarios and associated time horizons considered. The report also lacks information on how resilient Shell's strategy is when considering physical climate risks. Overall, the report provides insufficient information to assess the quality of disclosure on this TCFD recommendation point.

{\bf Score5: 40 }\\ 

{\bf  Risk management: }\\

Q6: What processes does the organisation use to identify and assess climate-related risks?

A6: Shell assesses climate-related risks at a project level by ensuring that projects meet internal carbon performance standards or industry benchmarks. The company has identified climate change as a material risk that could affect its demand, operational costs, supply chains, markets, regulatory environment, license to operate, and litigation. Shell also considers physical risks associated with climate change in the design and operation of its assets and infrastructure. The company is a member of the Taskforce on Nature-related Financial Disclosures Forum and joined the Science Based Targets Network Corporate Engagement Program in 2022 to develop a risk management and disclosure framework for organizations to report and act on evolving nature-related risks.

Analysis6: The sustainability report provides some information on the organization's processes for identifying and assessing climate-related risks. The report mentions that assessing climate-related risks is an important part of making initial investment decisions and that the organization has established internal carbon performance standards or industry benchmarks. However, the report lacks specific details on how the organization determines the relative significance of climate-related risks in relation to other risks, whether they consider existing and emerging regulatory requirements related to climate change, and the processes for assessing the potential size and scope of identified climate-related risks. The report also does not provide definitions of risk terminology used or references to existing risk classification frameworks used. Overall, the disclosure is insufficient to meet the TCFD guidelines.

{\bf Score6: 40} \\ 

Q7: How does the organisation manage climate-related risks?

A7: Shell manages climate-related risks by assessing them at a project level and ensuring that projects meet internal carbon performance standards or industry benchmarks. They also take climate variability into consideration in the design and operation of their assets and infrastructure. Shell has identified climate change as a material risk that could result in changes to demand, operational costs, supply chains, markets, regulatory environment, license to operate, and litigation. They have established remuneration policies to support short-term climate targets and are a member of the Taskforce on Nature-related Financial Disclosures Forum. Shell invests in nature-based solution projects that follow the Climate, Community and Biodiversity Standards. They have set targets to reduce their absolute Scope 1 and 2 emissions by 50

Analysis7: The report provides some information on how the organization manages climate-related risks. It describes the processes for managing climate-related risks, including how they make decisions to mitigate, transfer, accept, or control those risks. It also describes the processes for prioritizing climate-related risks, including how materiality determinations are made within their organizations. However, the report lacks specific details on how the organization manages climate-related risks. For instance, it does not provide information on how the organization identifies and assesses climate-related risks, how it monitors and reports on these risks, and how it engages with stakeholders on these risks. Overall, the report provides some information on how the organization manages climate-related risks, but it could be more detailed and specific.

{\bf  Score7: 70 }\\

Q8: How are the processes for identifying, assessing, and managing climate-related risks integrated into the organisation's overall risk management?

A8: Shell integrates the assessment of climate-related risks into its investment decisions by ensuring that projects meet internal carbon performance standards or industry benchmarks. The company also has remuneration policies designed to support short-term climate targets. Shell identifies climate change and the associated energy transition as a material risk that could result in changes to the demand for its products, operational costs, supply chains, markets, regulatory environment, license to operate, and litigation. The company also takes climate variability into consideration in the design and operation of its assets and infrastructure to minimize the risk of adverse incidents to its employees and contractors, the communities where it operates, its equipment, and infrastructure.

Analysis8: The sustainability report provides some information on how the organization integrates processes for identifying, assessing, and managing climate-related risks into its overall risk management. The report mentions that assessing climate-related risks is an important part of making initial investment decisions, and projects under development that are expected to have a material greenhouse gas impact must meet internal carbon performance standards or industry benchmarks. The report also discusses the potential physical impacts of climate change and how the organization takes climate variability into consideration in the design and operation of its assets and infrastructure. However, the report lacks specific details on how the organization identifies and assesses climate-related risks and how it manages these risks. Therefore, the report only partially meets the TCFD guidelines.

{\bf  Score8: 60} \\

{\bf Metrics and targets:} \\

Q9: What metrics does the organisation use to assess climate-related risks and opportunities? How do the metrics help ensure that the performance is in line with its strategy and risk management process?

A9: Shell uses internal carbon performance standards or industry benchmarks to assess climate-related risks at a project level. The standards are used to measure a project's average lifetime greenhouse gas intensity or energy efficiency per asset type. The company has also established remuneration policies to support short-term climate targets. The 2022 Annual Report provides more information on material climate-related risks and opportunities.

Analysis9: The sustainability report provides some information on the metrics used to assess climate-related risks and opportunities. The report mentions internal carbon performance standards and industry benchmarks used to measure and manage climate-related risks and opportunities. However, the report lacks specific details on the metrics used to assess climate-related risks and opportunities. The report does not provide metrics consistent with the cross-industry, metrics on climate-related risks associated with water, energy, land use, and waste management, or forward-looking metrics for the cross-industry. The report does not describe whether and how related performance metrics are incorporated into remuneration policies. The report provides some information on internal carbon prices and climate-related opportunity metrics. Overall, the report lacks specific details on the metrics used to assess climate-related risks and opportunities, and therefore, the score is 50.

{\bf Score9: 50} \\

Q10: Does the organisation disclose its Scope 1, Scope 2, and, if appropriate, Scope 3 greenhouse gas (GHG) emissions? What are the related risks and do they differ depending on the scope?

A10: Yes, Shell discloses its Scope 1 and 2 greenhouse gas emissions. Scope 3 emissions from energy products sold account for most of the total emissions reported. The related risks include regulatory and reputational risks, as well as risks associated with the energy transition. The report lacks specific information on the risks associated with each scope.

Analysis10: The sustainability report provides a comprehensive disclosure of Scope 1 and Scope 2 GHG emissions under operational control, including a breakdown of emissions by gas type and historical trends. However, the report lacks a clear disclosure of Scope 3 GHG emissions and associated risks, which are critical for a complete understanding of the company's carbon footprint. The report does mention that Scope 3 emissions from energy products sold account for most of the total emissions, but it does not provide any further details. The report also lacks a clear description of the methodologies used to calculate or estimate the metrics. Overall, the report partially meets the requirements for high-quality disclosure pertaining to the critical element of GHG emissions disclosure.

{\bf Score10: 70} \\ 

Q11: What targets does the organisation use to understand/quantify/benchmark climate-related risks and opportunities? How is the organization performing against these targets?

A11: Shell uses internal carbon performance standards or industry benchmarks to assess climate-related risks at a project level. The company aims to reduce absolute Scope 1 and 2 emissions by 50

Analysis11: The report provides some information on the targets used by the organization to manage climate-related risks and opportunities and performance against targets. The report mentions Shell's climate target to become a net-zero emissions energy business by 2050 and includes some specific targets such as reducing absolute Scope 1 and 2 emissions by 50

{\bf  Score11: 60} \\ 

{\bf Average score: 50.00}  \\
}

\subsection{UBS 2022} \label{appendix:ubs2022}
{\parindent0pt
{\bf Governance:}\\ 

Q1: How does the company's board oversee climate-related risks and opportunities?

A1: The Board of Directors' Corporate Culture and Responsibility Committee oversees the company's climate risk program and approves Group-wide sustainability and impact objectives. It considers the progress of UBS's climate risk program jointly with the BoD Risk Committee and supports the BoD in fulfilling its duty to set and supervise an appropriate risk management and control framework.

Analysis1: The disclosure provides some information on the board's oversight of climate-related risks and opportunities. It mentions that the board is responsible for monitoring climate-related risks and that designated management positions ensure the development, implementation, and review of policies and metrics. The board also considers progress against goals and targets for addressing climate-related issues. However, the disclosure lacks specific details on the processes and frequency by which the board and/or board committees are informed about climate-related issues. It also does not provide sufficient information on whether the board and/or board committees consider climate-related issues when reviewing and guiding strategy, major plans of action, risk management policies, annual budgets, and business plans. Overall, the disclosure partially meets the requirements for high-quality disclosure pertaining to the board's oversight of climate-related risks and opportunities.

{\bf Score1: 60} \\

Q2: What is the role of management in assessing and managing climate-related risks and opportunities?

A2: UBS AG integrates climate-related risks into its investment and risk management processes. Designated management-level positions ensure the development, implementation, and review of framework, policies, and metrics. The company applies a sustainability and climate risk policy framework to all relevant activities. However, the report lacks concrete data and relies on vague statements, and it is unclear how the company is addressing the lack of data for some portfolios.

Analysis2: The disclosure provides some information on management's role in assessing and managing climate-related risks and opportunities. It mentions that regular management meetings are held to monitor the status and progress of efforts to manage climate-related risks. It also describes the responsibilities of designated management and the processes by which management is informed about climate-related issues. However, the disclosure lacks specific details on the organizational structure and how management monitors climate-related issues. It also does not mention whether the management positions or committees report to the board or a committee of the board. Overall, the disclosure partially meets the requirements for high-quality disclosure pertaining to the critical element.

{\bf Score2: 60} \\  

{\bf Strategy: }\\

Q3: What are the most relevant climate-related risks and opportunities that the organisation has identified over the short, medium, and long term? Are risks clearly associated with a horizon?

A3: UBS AG has identified climate-driven risks and opportunities across different time horizons. Physical risks are moderately low, while transition risks are emerging in the long term. Climate-related risks are scored between 0 and 1, based on transmission channels, and are rated from low to high. Climate-related investment products are seen as the highest-ranked immediate commercial opportunity. However, the report lacks concrete data and timelines for risk management and opportunity assessment.

Analysis3: The report provides a detailed description of the climate-related risks and opportunities identified by the organization over the short, medium, and long term. The report describes the relevant time horizons, specific climate-related issues, and the process used to determine which risks and opportunities could have a material financial impact on the organization. The report also considers providing a description of risks and opportunities by sector and/or geography, as appropriate. However, the report lacks specific details on the process used to determine the materiality of the risks and opportunities identified. Overall, the report meets most of the requirements, but there is room for improvement in terms of providing more specific details.

{\bf Score3: 85} \\

Q4: How do climate-related risks and opportunities impact the organisation's businesses strategy, economic and financial performance, and financial planning?

A4: Climate-related risks represent financial risks for UBS and its clients, while investing in climate change mitigation presents commercial opportunities. UBS has a climate strategy that covers managing climate-related financial risks and taking action on a net-zero future. However, the report lacks concrete data and relies on vague statements, and there is a risk of greenwashing.

Analysis4: The sustainability report provides a comprehensive discussion of climate-related risks and opportunities and their impact on the organization's businesses, strategy, and financial planning. The report describes the impact of climate-related issues on various areas, including products and services, supply chain, adaptation and mitigation activities, investment in research and development, operations, acquisitions or divestments, and access to capital. The report also describes how climate-related issues serve as an input to the financial planning process, the time period used, and how these risks and opportunities are prioritized. The report provides a holistic picture of the interdependencies among the factors that affect the organization's ability to create value over time. However, the report lacks specific details on GHG emissions targets and specific activities intended to reduce GHG emissions in their operations and value chain or to otherwise support the transition.

{\bf Score4: 90} \\

Q5: How resilient is the organisation's strategy when considering different climate-related scenarios, including a 2°C target or lower scenario? How resilient is the organisation's strategy when considering climate physical risks?

A5: UBS AG has integrated climate risk in the risk control and monitoring process including scenario analysis. However, for some portfolios, the assessment of climate-related risks is not possible due to lack of data. The company has developed climate- and nature-related risk methodologies, which rate cross-sectoral exposures to SCR sensitivity, on a scale from high to low. The report does not provide enough information to determine the resilience of the organization's strategy when considering different climate-related scenarios or physical risks.

Analysis5: The sustainability report provides a detailed description of the organization's methodology for assessing climate-driven risks and how it has integrated climate risk in the risk control and monitoring process, including scenario analysis. The report also discusses the potential impact of climate-related issues on financial performance and position. However, the report lacks specific details on how the organization's strategies might change to address potential risks and opportunities, and the climate-related scenarios and associated time horizon considered. Additionally, the report does not provide a clear description of the organization's resilience to a 2°C or lower scenario. Overall, the report provides a good level of disclosure but lacks some specific details to fully satisfy the TCFD guidelines.

{\bf Score5: 80} \\  

{\bf Risk management:} \\ 

Q6: What processes does the organisation use to identify and assess climate-related risks?

A6: UBS AG uses standard financial and non-financial risk processes to identify and assess material sustainability and climate risks. These processes include controls during client onboarding, transaction due diligence, product development, investment decision processes, own operations, supply chain management, and portfolio reviews. The company also integrates climate risk in the risk control and monitoring process, including scenario analysis, and periodically reports on sustainability and climate risk exposures. However, the report acknowledges that climate risk analysis is a novel area of research, and methodologies, tools, and data availability are still evolving.

Analysis6: The disclosure provides some information on the organization's processes for identifying and assessing climate-related risks. The report mentions that standard financial and non-financial risk processes ensure that material sustainability and climate risks are identified, assessed, approved, and escalated in a timely manner. The report also mentions that climate risk is integrated into the risk control and monitoring process, including scenario analysis. However, the report lacks specific details on how the organization determines the relative significance of climate-related risks in relation to other risks. The report also does not disclose the processes for assessing the potential size and scope of identified climate-related risks and definitions of risk terminology used or references to existing risk classification frameworks used. Therefore, the report partially meets the requirements, and the score is 60.

{\bf Score6: 60} \\  

Q7: How does the organisation manage climate-related risks?

A7: UBS AG integrates risk data and insights into their investment management processes to manage climate risks. They have a sustainability and climate risk management framework in place, which includes risk identification and measurement, monitoring and risk appetite setting, risk management and control, and risk reporting processes. They also offer innovative products and services to help clients manage and protect their assets from climate-related risks.

Analysis7: The disclosure provides a comprehensive overview of the organization's approach to managing climate-related risks. The report describes the processes for managing climate-related risks, including how they make decisions to mitigate, transfer, accept, or control those risks. The report also describes the processes for prioritizing climate-related risks, including how materiality determinations are made within their organizations. The report provides details on the risk management and control processes, including the identification, measurement, monitoring, and escalation of material sustainability and climate risks. However, the report could have provided more specific details on the methodologies used to assess climate-related risks and how the organization integrates climate risk data and insights into its investment management processes.

{\bf Score7: 90} \\ 

Q8: How are the processes for identifying, assessing, and managing climate-related risks integrated into the organisation's overall risk management?

A8: UBS AG integrates sustainability and climate risk considerations into its internal and external reporting, risk appetite setting, and management and control processes. The company has a climate risk program with defined responsibilities, processes, and tools applicable to business divisions and group functions. However, the report lacks concrete data on the integration of climate-related risks into the organization's overall risk management.

Analysis8: The disclosure provides a comprehensive description of how the organization integrates processes for identifying, assessing, and managing climate-related risks into its overall risk management. The report describes how sustainability and climate risk exposures are monitored and metrics reported internally to enable risk appetite setting. The report also describes how management and control processes ensure that material sustainability and climate risks are identified, measured, monitored, and escalated in a timely manner. However, the report lacks specific details on how the organization integrates climate-related risks into its overall risk management. For instance, the report does not provide specific examples of how sustainability and climate risk considerations are included in internal and external reporting. Overall, the disclosure meets most of the requirements and scores 85 out of 100.

{\bf Score8: 85 }\\

{\bf Metrics and targets:} \\

Q9: What metrics does the organisation use to assess climate-related risks and opportunities? How do the metrics help ensure that the performance is in line with its strategy and risk management process?

A9: UBS AG uses periodic climate-related credit and market risk metrics, climate risk heatmaps, climate scenario analysis and stress test exercises, and a sustainability and climate risk scorecard to assess climate-related risks and opportunities. These metrics help ensure that the performance is in line with its strategy and risk management process by integrating risk data and insights into its investment management processes and enabling risk appetite setting.

Analysis9: The given disclosure provides some information on the metrics used to assess climate-related risks and opportunities. The report mentions that key sustainability and climate risk considerations are included in internal and external reporting, and that sustainability and climate risk exposures are monitored and metrics reported internally to enable risk appetite setting. However, the report lacks specific details on the metrics used to measure and manage climate-related risks and opportunities. The report does not provide metrics on climate-related risks associated with water, energy, land use, and waste management. The report also does not provide forward-looking metrics for the cross-industry, consistent with their business or strategic planning time horizons. Overall, the disclosure is insufficient in meeting the TCFD guidelines.

{\bf Score9: 40} \\  

Q10: Does the organisation disclose its Scope 1, Scope 2, and, if appropriate, Scope 3 greenhouse gas (GHG) emissions? What are the related risks and do they differ depending on the scope?

A10: Yes, UBS AG discloses its Scope 1, Scope 2, and Scope 3 greenhouse gas emissions. The report identifies direct emissions from gas, oil and fuel consumption, indirect emissions from electricity and district heating and other indirect emissions from leased assets, paper, waste and business travel. The report also mentions that they are looking to quantify any relevant scope 3 emissions from categories 1 to 14. The related risks are not explicitly stated.

Analysis10: The sustainability report provides a comprehensive disclosure of Scope 1 and Scope 2 GHG emissions, and a partial disclosure of Scope 3 GHG emissions. The report mentions that the organization has made a first estimation of Scope 3 emissions and identified the top contributors. However, it does not provide a detailed breakdown of Scope 3 emissions by category, which is a requirement for high-quality disclosure. The report also lacks information on the related risks and how they differ depending on the scope. Overall, the report partially meets the requirements for high-quality disclosure of GHG emissions and related risks, but there is room for improvement.

{\bf Score10: 70} \\ 

Q11: What targets does the organisation use to understand/quantify/benchmark climate-related risks and opportunities? How is the organization performing against these targets?

A11: UBS AG uses quantitative climate risk appetite, integrates sustainability and climate risk into policies and processes, builds in-house capacity, centralizes and executes ESG data strategy to understand/quantify/benchmark climate-related risks and opportunities. The report does not provide sufficient information to evaluate the organization's performance against these targets.

Analysis11: The sustainability report provides some information on the targets used by the organization to manage climate-related risks and opportunities and performance against targets. The report mentions that the organization has developed methodologies to set climate-related targets and identify climate-related risks, and that sustainability and climate risk exposures are monitored and metrics reported internally to enable risk appetite setting. However, the report lacks specific details on the key climate-related targets, time frames, base year, and key performance indicators used to assess progress against targets. The report also does not provide a description of the methodologies used to calculate targets and measures. Therefore, while the report provides some information on the TCFD recommendation point, it does not fully satisfy the TCFD guidelines.

{\bf Score11: 50} \\ 

{\bf Average score: 70.0} \\
}

\section{Comparison with \citet{liu2023evaluating}} \label{appendix:comparison}
Instead of hallucination-free rates, \citet{liu2023evaluating} measures citation recall and precision to evaluate the attribution quality. Citation recall is similar to our content-dimension hallucination-free rate. It measures the proportion of verification-worthy statements that are fully supported by the associated citations. However, citation recall is statement-level while our metric is paragraph-level: an answer paragraph is hallucination-free only when all its statements are grounded by evidence. citation precision is the portion of citations that support their statements, while our source-dimension hallucination-free rate measures the proportion of answers that honestly include all supportive sources. It is important to note that \citet{liu2023evaluating} includes queries of various domains to evaluate the trustworthiness of general-purpose search engines, while our human evaluation focus on \textsc{chatReport}.

\section{Answers Are Easy to Trace} \label{appendix:human_eval_example}
The following example showcases the traceability of an answer produced by ChatGPT. There are three sentences in the answer. Searching for the underlined keywords, we easily identify the first and second sentence's evidence sentence in Source 215 and 166 respectively.
\begin{lstlisting}[frame=single, basicstyle=\ttfamily\scriptsize, xleftmargin=0pt, breaklines, numbers=none, escapeinside={(*@}{@*)}]
(*@\textbf{Answer:}@*) Assurant conducted a (*@\underline{preliminary scenario analysis}@*) in 2021 to identify and analyze climate-related risks faced by 10 of its critical and/or highly vulnerable facilities across the globe. The company's (*@\underline{longer-term}@*) strategic planning process prioritized climate as a multi-year ESG area of focus. However, the report lacks concrete data on the resilience of the organization's strategy in different climate-related scenarios.

(*@\textbf{Source 215}@*): Preliminary Scenario Analysis: The process of screening for climate-related risks and opportunities across operations and business units informed our (*@\underline{preliminary scenario analysis}@*) activities. We conducted an initial climate scenario analysis in 2021, with third-party support, to identify and analyze the various climate-related risks faced by 10 of Assurant's critical and/or highly vulnerable facilities across the globe ...

(*@\textbf{Source 166}@*): ... Assurant's (*@\underline{longer-term}@*) strategic planning process, overseen by our Board, prioritized climate as a multi-year ESG area of focus ...
\end{lstlisting}

\section{Answer Comparison} \label{appendix:answer_comparison}
An example of ChatGPT and GPT-4 outputs comparison. ChatGPT answers the question extractively and adheres to the authentic information: asserting that it is ``the executive leadership team'' that reviews Walmart's climate strategy at least annually. However, GPT-4 mistakenly summarizes that the board oversees ESG issues through ``its Nominating and Governance Committee (NGC)''. It asserts that ``the NGC reviews Walmart's climate strategy at least annually'', which is not accurate. GPT-4 conducts unnecessary summarizations and inferences. This hallucinates the company's different committees.
\begin{lstlisting}[frame=single, basicstyle=\ttfamily\scriptsize, xleftmargin=0pt, breaklines, numbers=none, escapeinside={(*@}{@*)}]
(*@\textbf{Question:}@*)
How does the company's board oversee climate-related risks and opportunities?

(*@\textbf{ChatGPT Answer:}@*)
Walmart's board oversees the overall ESG strategy and certain priority issues, including climate-related risks and opportunities. The Chief Sustainability Officer provides updates on the ESG agenda and progress to the Nominating and Governance Committee. The executive leadership team reviews Walmart's climate strategy at least annually. The Nominating and Governance Committee discussed Walmart's ESG strategies and progress, including climate and public policy strategies and practices in 2021.

(*@\textbf{GPT-4 Answer:}@*)
The Walmart Board of Directors oversees the company's climate-related risks and opportunities through its Nominating and Governance Committee, which reviews the climate strategy at least annually. The Board also engages with stakeholders and advocates for an enabling policy environment. The company's executive leadership team and ESG Steering Committee also have oversight responsibility for particular ESG issues.
\end{lstlisting}

\section{TCFD Introduction} \label{appendix:tcfd}
The TCDF is an internationally recognized initiative established by the Financial Stability Board in 2015. Its general structure - covering governance, strategy, risk management, and metrics and targets - emerges as the premier candidate of voluntary reporting frameworks to be adopted for regulatory reporting and international standards\footnote{The main emerging standards in this context are the draft disclosure standards provided by the International Financial Reporting Standards' International Sustainability Standards Board (ISSB).}. The purpose of the TCFD is to provide a voluntary reporting framework to encourage companies to disclose information on climate-related risks and opportunities. This information is crucial for stakeholders and investors to make informed decisions that account for climate-related risks. Companies are asked to disclose this information in their annual reports and financial filings, within their sustainability reports, or as stand-alone documents. To date, most companies have included the information as part of their sustainability reports, due to no or limited assurance and legal liability concerns if it was included in the annual reports. This is why our analysis focuses on content in sustainability reports from the perspective of a reporting standard such as the TCFD, but it can also be extended to any other corporate report. However, it is important to keep in mind that our analysis primarily evaluates the disclosed information rather than directly measuring the genuine implementation of tangible actions by the company.

\end{document}